\documentclass[
 twocolumn,
 superscriptaddress,
 nofootinbib,
 amsmath,
 amssymb,
 aps,
]{revtex4}

\usepackage{graphicx} 
\usepackage{dcolumn} 
\usepackage{bm} 

\usepackage{hyperref} 

\usepackage{xcolor}
\allowdisplaybreaks

\usepackage{mathtools}

\usepackage[whole]{bxcjkjatype}
\usepackage[linesnumbered, ruled, vlined]{algorithm2e}

\makeatletter
\newcommand{\vast}{\bBigg@{3.0}}
\newcommand{\Vast}{\bBigg@{4.0}}

\makeatother

\usepackage{coffeestains}

\usepackage{xr}
\renewcommand{\theequation}{\arabic{section}.\arabic{equation}}
\makeatletter
\@addtoreset{equation}{section}
\makeatother

\usepackage{placeins}

\begin{document}

\preprint{APS/123-QED}

\title{Phase transition on a context-sensitive random language model with short range interactions}

\author{Yuma Toji}

\affiliation{
	Graduate School of Information Science and Technology,
	Hokkaido University, Sapporo, Hokkaido 060-0814, Japan
}


\author{Jun Takahashi}

\affiliation{
	Institute for Solid State Physics, The University of Tokyo, 5-1-5 Kashiwanoha, Kashiwa, Chiba 277-8581, Japan
}


\author{Vwani Roychowdhury}


\affiliation{
	Henry Samueli School of Engineering and Applied Science, \\
	University of California, Los Angeles, California 90095
}

\author{Hideyuki Miyahara}

\email{miyahara@ist.hokudai.ac.jp, hmiyahara512@gmail.com}

\thanks{corresponding author.}

\affiliation{
	Graduate School of Information Science and Technology,
	Hokkaido University, Sapporo, Hokkaido 060-0814, Japan
}

\date{\today}

\begin{abstract}
	Since the random language model was proposed by E. DeGiuli [Phys. Rev. Lett. 122, 128301], language models have been investigated intensively from the viewpoint of statistical mechanics. Recently, the existence of a Berezinskii--Kosterlitz--Thouless transition was numerically demonstrated in models with long-range interactions between symbols.
	In statistical mechanics, it has long been known that long-range interactions can induce phase transitions. Therefore, it has remained unclear whether phase transitions observed in language models originate from genuinely linguistic properties that are absent in conventional spin models.
	In this study, we construct a random language model with short-range interactions and numerically investigate its statistical properties. Our model belongs to the class of context-sensitive grammars in the Chomsky hierarchy and allows explicit reference to contexts.
	We find that a phase transition occurs even when the model refers only to contexts whose length remains constant with respect to the sentence length. This result indicates that finite-temperature phase transitions in language models are genuinely induced by the intrinsic nature of language, rather than by long-range interactions.
\end{abstract}


\maketitle


\section{Introduction}

Phase transitions, such as the liquid-gas transition, metal-insulator transition, magnetic phase transition, and superconducting transition, are ubiquitous phenomena in statistical mechanics, arising from collective behavior in many-body systems~\cite{Huang1987,Goldenfeld1992,Landau1980}.
Most systems around us can be regarded as many-body systems because they involve interactions among their constituent elements.
It is therefore natural to expect that phase-transition-like phenomena may also appear beyond traditional physical systems.
In recent years, phenomena analogous to phase transitions have been reported in natural language processing and have attracted considerable attention, suggesting a deep connection between linguistic structure and statistical physics.
In particular, emergent abilities and scaling behaviors in large language models (LLMs)~\cite{Kaplan_001,Wei_001} have been interpreted as manifestations of such collective effects.
Notably, prior to these striking findings, E.~DeGiuli formulated the random language model, which describes language generation as a statistical ensemble over grammars, and argued for the existence of phase transitions in language models~\cite{De-Giuli_001}, unfortunately leading to a dispute~\cite{Nakaishi_001}.
In this framework, sentences are generated probabilistically according to randomly drawn grammar rules, enabling linguistic structure to be analyzed using tools from statistical mechanics.
This model establishes a concrete bridge between formal language theory and many-body physics, opening the possibility of studying linguistic phenomena in terms of thermodynamic phases and critical behavior.

Building upon this framework, we previously investigated a variant of the random language model incorporating long-range interactions between symbols~\cite{toji2026berezinskii}.
Through numerical simulations, we demonstrated the existence of a Berezinskii--Kosterlitz--Thouless (BKT) transition~\cite{Berezinskii_001, Berezinskii_002, Kosterlitz_001, Kosterlitz_004, Kosterlitz_002} in this system, revealing nontrivial critical behavior analogous to that observed in low-dimensional statistical-mechanical models.
This result provided strong evidence that language models can exhibit genuine phase transitions.
However, long-range interactions are well known to induce phase transitions even in otherwise simple spin systems~\cite{Dyson_001, Dyson_002, Dyson_003, Tomita_001, Martinez-Herrera_001, Kosterlitz_004}.
Therefore, it remains unclear whether the observed phase transitions in language models genuinely reflect intrinsic properties of language, or whether they are merely consequences of long-range couplings.

In the present study, we address this question by constructing a random language model that incorporates only short-range interactions.
Specifically, we introduce a generative model in analogy with the one-dimensional Potts model, where interactions are restricted to finite-range contexts whose length remains constant with respect to the sentence length~\cite{Wu_001}.
Sentences are generated from this model and analyzed using standard statistical-mechanical observables, including magnetization, susceptibility, Binder cumulant, and correlation functions.
Our numerical analysis reveals that a phase transition occurs even in the absence of long-range interactions.
Remarkably, the transition is identified as a BKT transition as in the previously studied long-range model.
In this transition, correlation functions exhibit power-law decay not only at the critical point but throughout an extended critical phase below it, in contrast to conventional first- or second-order phase transitions.
These findings demonstrate that nontrivial critical phenomena in language models do not rely on long-range couplings, but instead emerge from structural properties intrinsic to the linguistic generative process.

This paper is organized as follows.
In Sec.~\ref{main_sec_model_001_001}, we explain the details of the proposed model.
The physical quantities calculated from the model are described in Sec.~\ref{main_sec_method_001_001}.
The results of the representative parameters are shown in Sec.~\ref{main_sec_numerical_simulation_001_001}, and discussed in Sec.~\ref{main_sec_discussion_001_001}.
Finally, we conclude this paper in Sec.~\ref{main_sec_conclusion_001_001}.

\section{Proposed model} \label{main_sec_model_001_001}

In this paper, we consider the model that has the following three generative rules~\cite{harrison1978introduction, rozenberg2012handbook, linz2022introduction, meduna2014formal, Toji_001}:
\begin{subequations} \label{main_eq_rule_003_001}
	\begin{align}
		X                   & \to x                   & (\text{probability} & : qt), \label{main_eq_rule_004_001}        \\
		X                   & \to YZ                  & (\text{probability} & : q (1 - t)), \label{main_eq_rule_004_002} \\
		Z_- X Z_+ & \to Z_- Y Z_+ & (\text{probability} & : (1 - q)), \label{main_eq_rule_004_003}
	\end{align}
\end{subequations}
where the lowercase and uppercase letters represent terminal and non-terminal symbols, respectively.

We elaborate on the definitions of three rules in Eq.~\eqref{main_eq_rule_003_001}, respectively.
The rule~\eqref{main_eq_rule_004_001} transforms a non-terminal symbol into its \textit{corresponding} terminal symbol.
Here, \textit{corresponding} means that there is a one-to-one relation between a non-terminal symbol $A_k$ and a terminal symbol $a_k$, so a production rule like $A_2 \to a_4$ does not exist.
Thus, the number of terminal and non-terminal symbols are specified by $K$; that is, the model handles $K$ types of terminal symbols $a_1,a_2,\ldots,a_K$, and $K$ types of non-terminal symbols $A_1,A_2,\ldots,A_K$.

The rule~\eqref{main_eq_rule_004_002} transforms a non-terminal symbol into two non-terminal symbols; in other words, this rule increases the number of non-terminal symbols in the sentence.
This rule has a parameter $\epsilon \in [0,1]$, which controls the tendency of what symbols appear on the right-hand side of the rule~\eqref{main_eq_rule_004_002}.
Specifically, when the symbol on the left-hand side of the rule~\eqref{main_eq_rule_004_002} is $A_k$, the symbols $Y$ and $Z$ in the right-hand side of the rule~\eqref{main_eq_rule_004_002} are selected as follows:
\begin{align}
	Y,Z =
	\left\{
	\begin{array}{ccccc}
		A_k                              & (\text{probability} & : & 1-\frac{(K-1)\epsilon}{K} & ), \\
		A \in \Sigma_{\mathrm{nt}} \setminus \{A_k\} & (\text{probability} & : & \frac{\epsilon}{K}        & ).
	\end{array}
	\right.
\end{align}
Note that $\Sigma_{\mathrm{nt}} = \{A_1,A_2,\ldots,A_K\}$ denotes the set of non-terminal symbols.
According to this formulation, $Y$ and $Z$ are always selected as $X$ when $\epsilon=0$, and selected uniformly when $\epsilon=1$.
The model that consists of only the rules~\eqref{main_eq_rule_004_001} and \eqref{main_eq_rule_004_002} corresponds to a context-free grammar (CFG).

The rule~\eqref{main_eq_rule_004_003} characterizes the context sensitivity of the model.
This rule replaces the symbol $X$ with the symbol $Y$, depending on the $Z_-$ and $Z_+$, the symbols before and after $X$.
If $X$ is $A_k$, the rule~\eqref{main_eq_rule_004_003} first selects a destination symbol $Y$ from the set of non-terminal symbols other than $A_k$.
Suppose $Y$ is $A_j$.
Then, the probability $p$ of applying the rule~\eqref{main_eq_rule_004_003} is calculated as follows:
\begin{subequations}
\begin{align}
	p        & = \min \bigg( 1, \exp \bigg( -\frac{\Delta E}{k_{\mathrm{B}} T} \bigg) \bigg), \\
	\Delta E & \coloneqq J (\delta_{\sigma_i,\sigma_{i - 1}} - \delta_{\tilde{\sigma}_i,\sigma_{i - 1}} + \delta_{\sigma_i,\sigma_{i + 1}} - \delta_{\tilde{\sigma}_i,\sigma_{i + 1}}).
\end{align}
\end{subequations}
Here, $k_{\mathrm{B}} T$ is the temperature parameter of the rule~\eqref{main_eq_rule_004_003} and the parameter $J$ is a positive constant.
The variable $\sigma_i \in \{ 1, 2, \dots, K \}$ denotes the $i$-th symbol in the sentence.
If the $i$-th symbol is $a_k$ or $A_k$, then $\sigma_i$ is equal to $k$.

This process is the same as the one used in the simulation of the $K$-state Potts model by the Metropolis-Hastings method with single-spin flips~\cite{Landau_001, Binney_001, Newman_001}.
In other words, this model adopts the interactions of the Potts model as interactions between symbols in the sentences.
When $T$ is small, replacements based on the contexts tend to occur, whereas when $T$ is large, random replacements tend to occur.

\begin{figure}[ht]
	\centering
	\includegraphics[scale=0.60]{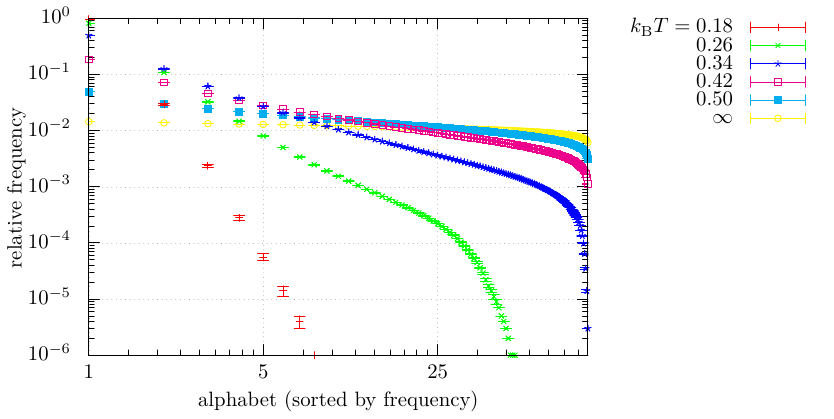}
	\includegraphics[scale=0.60]{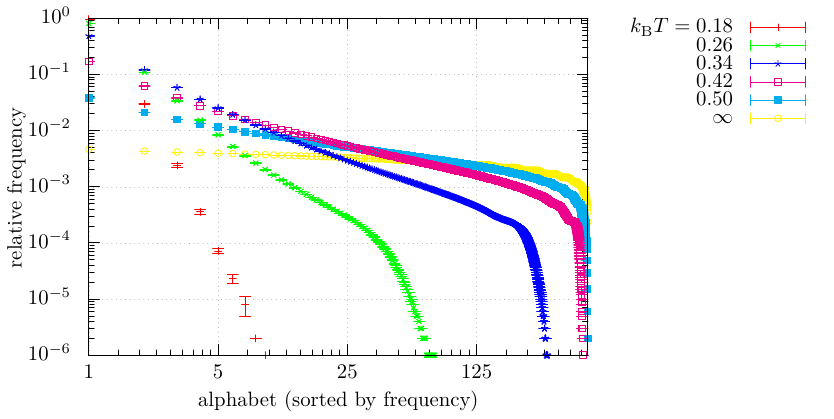}
	\caption{Ranking of symbol appearance frequencies for (upper) $K=100$ and (lower) $K=500$, plotted on log--log scales. The critical temperature in both cases is $k_{\mathrm{B}} T \simeq 0.42$. We set $J = 1.0$, $q = 10^{-1.0}$, $t = 0.0$, and $\epsilon = 0.00$. The curves for different $k_{\mathrm{B}} T$ are overlaid.}
	\label{main_fig_zipfs_law_proposed_model}
\end{figure}

Our model is a highly simplified model of language.
Nevertheless, when we increase $K$, the number of possible states for each symbol, we confirm that the model reproduces phenomena similar to Zipf's law observed in natural languages.
Figure~\ref{main_fig_zipfs_law_proposed_model} shows the relative frequencies of symbols, ranked in descending order and plotted on log--log axes,
for $K=100$ and $K=500$.
Although the model is based on the strong assumption that all symbols are treated completely symmetrically, the fact that it can reproduce Zipf-like behavior provides supporting evidence for the validity of the results obtained from this model.

From the viewpoint of formal language theory, the present model belongs to the class of context-sensitive grammars in the Chomsky hierarchy, whereas the random language model introduced by E. DeGiuli corresponds to a context-free grammar~\cite{Jager_001}.
In this sense, our construction can be regarded as a natural extension of DeGiuli's framework, incorporating explicit contextual dependence while remaining within a probabilistic generative setting.

\section{Physical quantities} \label{main_sec_method_001_001}

\begin{figure}[t]
	\centering
	\includegraphics[scale=0.18]{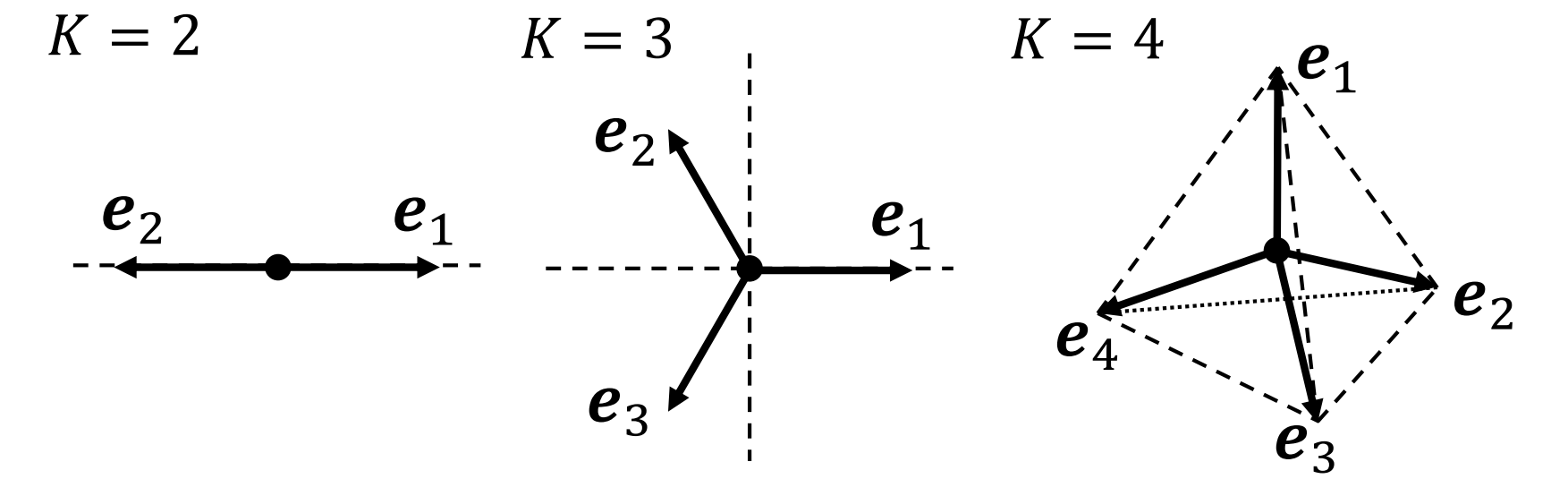}
	\caption{A schematic diagram of the vectors $\bm{e}_{k}$ for $K=2,3,$ and $4$.
		They sum to the zero vector,
		and the inner products of any two different vectors are constant.}
	\label{main_fig_symplectic_vector_001_001}
\end{figure}

In this section, we describe the physical quantities calculated from the model explained in Sec.~\ref{main_sec_model_001_001}.
As an order parameter, we used the magnetization $M$ shown in Eq.~\eqref{main_eq_magnetization_001_001} that is defined in the same way as in the Potts model~\cite{Wu_001}:
\begin{align}
	M & \coloneqq \|\bm{M}\|, \label{main_eq_magnetization_001_001} 
\end{align}
where
\begin{align}
	\bm{M} & \coloneqq \frac{1}{N}\sum_{i=1}^{N}\bm{e}_{\sigma_i}. \label{main_eq_magnetization_001_002}
\end{align}
Here, $N$ denotes the length of the sentence.
The vectors $\{ \bm{e}_{k} \}_{k=1}^K$ satisfy the following conditions:
\begin{subequations}
	\begin{align}
		\sum_{k=1}^{K} \bm{e}_k & = \bm{0}, \label{main_eq_symplectic_vector_001_001} \\
		\bm{e}_k \cdot \bm{e}_l & =
		\begin{cases}
			1             & (k = l),   \\
			\text{const.} & (k \ne l).
		\end{cases} \label{main_eq_symplectic_vector_001_002}
	\end{align}
\end{subequations}
Equation~\ref{main_eq_symplectic_vector_001_001} leads to $M = 0$ in Eq.~\eqref{main_eq_magnetization_001_001} for a paramagnetic state and Eq.~\eqref{main_eq_symplectic_vector_001_002} is a condition that reflects the system's symmetry in physical quantities.
The vectors $\bm{e}_k$ for $K=2,3,$ and $4$ are illustrated in Fig.~\ref{main_fig_symplectic_vector_001_001}.

The magnetic susceptibility $\chi$ is defined as the variance of the magnetization $M$:
\begin{align}
	\chi \coloneqq N(\langle M^2 \rangle - \langle M \rangle^2). \label{main_eq_susceptibility_001_001}
\end{align}
The susceptibility, Eq.~\eqref{main_eq_susceptibility_001_001}, is divergent at a critical point; thus it is quite important to see its behavior to discuss a phase transition via numerical simulations.

The Binder parameter $U$ is the kurtosis of the distribution of $M$, normalized such that $U$ equals $0$ when the distribution of the magnetization is Gaussian, and equals $1$ when the distribution is a delta function:
\begin{align}
	U \coloneqq -\frac{K - 1}{2} \left( \frac{\langle M^4 \rangle}{\langle M^2 \rangle^2} - \frac{K + 1}{K - 1} \right).
	\label{main_eq_Binder_parameter_001_001}
\end{align}
The Binder parameter plays an important role in distinguishing between the standard second-order phase transition and the BKT transition~\cite{Tuan_001, Hasenbusch_001}.

The correlation function is also an important quantity to discuss phase transition since it typically shows a power-law decay at a critical point.
We define it as follows~\cite{Lin_001}:
\begin{align}
	G_{k, l} \coloneqq \langle \bm{e}_{\sigma_k} \cdot \bm{e}_{\sigma_l} \rangle - \langle \bm{e}_{\sigma_k} \rangle \cdot \langle \bm{e}_{\sigma_l} \rangle.
	\label{main_eq_correlation_function_001_001}
\end{align}
To observe the power-law decay of the correlation function associated with critical phenomena, it is sufficient to evaluate the following quantity:
\begin{align}
	\tilde{G}_{k, l} \coloneqq \langle \bm{e}_{\sigma_k} \cdot \bm{e}_{\sigma_l} \rangle.
	\label{main_eq_correlation_function_001_002}
\end{align}
In this paper, we compute Eq.~\eqref{main_eq_correlation_function_001_002} instead of Eq.~\eqref{main_eq_correlation_function_001_001}.

Finally, we describe the finite-size scaling~\cite{Ardourel_001, Hsieh_001, Zuo_001, Ueda_001}.
\begin{align}
	\frac{\tilde{\chi}(T,N)}{N^{\frac{\gamma}{\nu}}} = \tilde{f}_{\tilde{\chi}}(N^{\frac{1}{\nu}}t).
	\label{main_eq_finite_size_scaling_function_001_001}
\end{align}
Here, $t \coloneqq (T - T_{\mathrm{c}}) / T_{\mathrm{c}}$ and $\tilde{\chi}$ is defined as
\begin{align}
	\tilde{\chi} \coloneqq N \langle M^2 \rangle.
	\label{main_eq_susceptibility_001_002}
\end{align}
The finite-size scaling relation holds around the critical temperature.
When we plot $\tilde{\chi}(T,N)/N^{\gamma/\nu}$ against $N^{1/\nu}t$ for different system sizes $N$, the data collapse onto a single universal curve $\tilde{f}$.

\section{Numerical simulations} \label{main_sec_numerical_simulation_001_001}

\begin{figure}[t]
	\centering
	\includegraphics[scale=0.60]{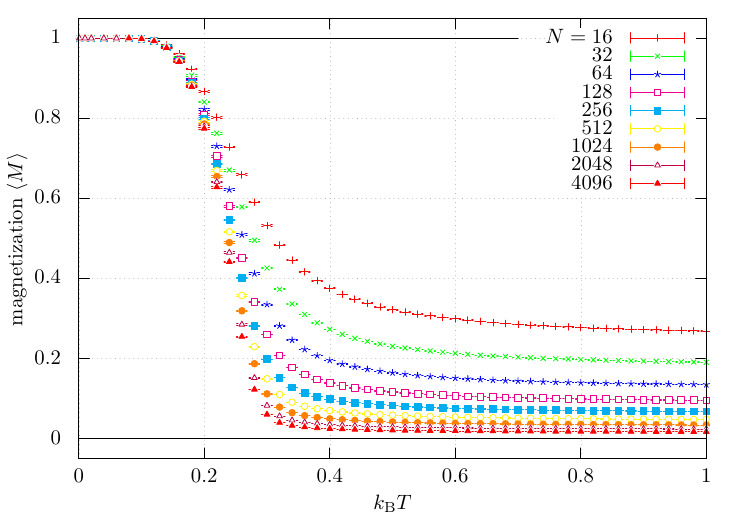}
	\includegraphics[scale=0.60]{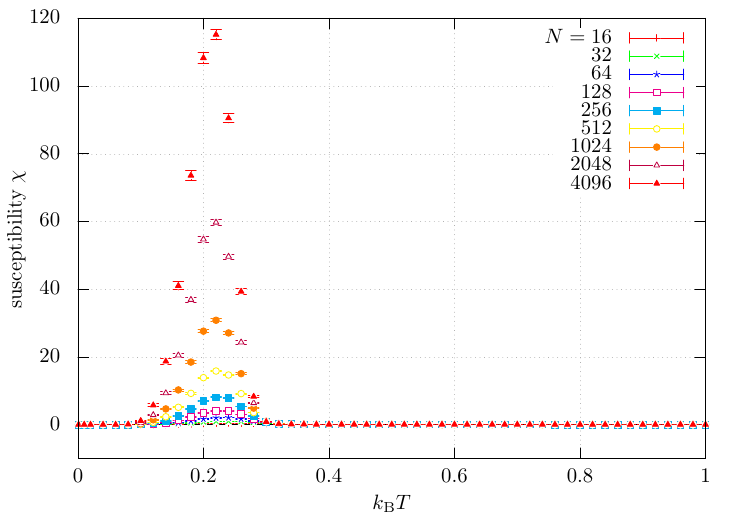}
	\includegraphics[scale=0.60]{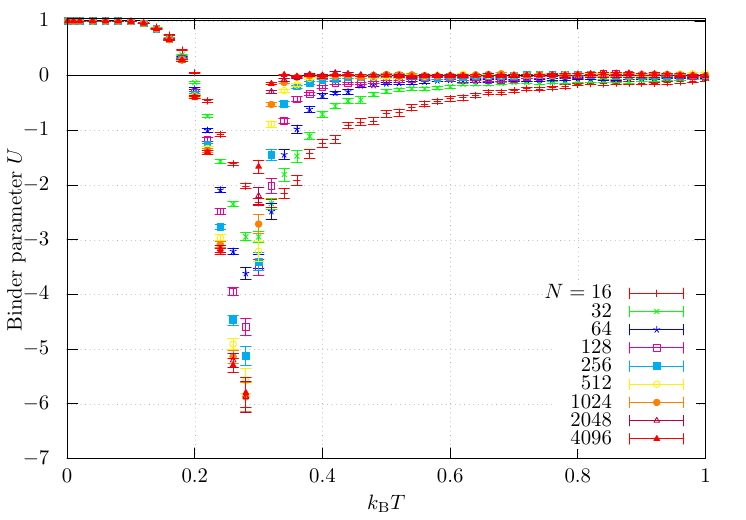}
	\caption{Temperature dependence of (upper) the magnetization, Eq.~\eqref{main_eq_magnetization_001_001}, (middle) the susceptibility, Eq.~\eqref{main_eq_susceptibility_001_001}, and (lower) the Binder parameter, Eq.~\eqref{main_eq_Binder_parameter_001_001}. We set $K = 20, J = 1.0, q = 10^{-2.0}, t = 0.0$, and $\epsilon = 0.00$. We show the results for various system sizes $N = 16, 32, \dots, 4096$, and the curves for different $N$ are overlaid in each panel.}
	\label{main_fig_mag_sus_Bin_K=20_R=1_q_-2.0_e=0.00_001_001}
\end{figure}

In this section, we present the results for $K = 20, \epsilon = 0.00$ as representative cases.
Figure~\ref{main_fig_mag_sus_Bin_K=20_R=1_q_-2.0_e=0.00_001_001} shows the magnetization, susceptibility and Binder parameter for $K = 20, t = 0.0, q = 10^{-2.0}$, and $\epsilon = 0.00$.
For $t = 0.0$, we can consider a simple thermodynamic limit, since the production rule~\eqref{main_eq_rule_004_001} is not chosen, and the sentence continues to extend indefinitely.
The magnetization and susceptibility show singularity and divergence, respectively, around $k_{\mathrm{B}} T \sim 0.24$.
The Binder parameter also changes from $0$ to negative values around $k_{\mathrm{B}} T \sim 0.24$, and then approaches $1$ as the temperature decreases.

\begin{figure}[t]
	\centering
	\includegraphics[scale=0.60]{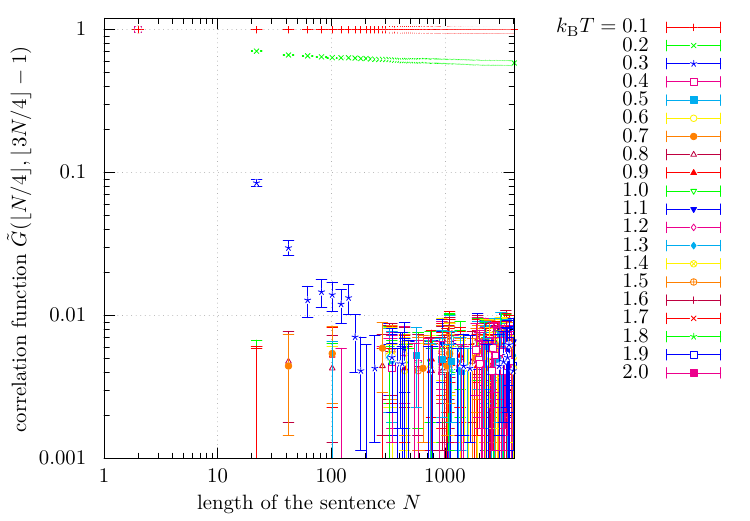}
	\caption{The correlation function, Eq.~\eqref{main_eq_correlation_function_001_002}, with $i = \lfloor N/4 \rfloor$ and $j = \lfloor 3N/4 \rfloor - 1$. We set $K = 20, J = 1.0, q = 10^{-2.0}, t = 0.0$, and $\epsilon = 0.00$. We show the results for various $k_{\mathrm{B}} T = 0.1, 0.2, \dots, 2.0$, and the curves for different $k_{\mathrm{B}} T$ are overlaid.}
	\label{main_fig_correlation_K=20_R=1_q_-2.0_e=0.00_001_001}
\end{figure}

\begin{figure}[t]
	\centering
	\includegraphics[scale=0.60]{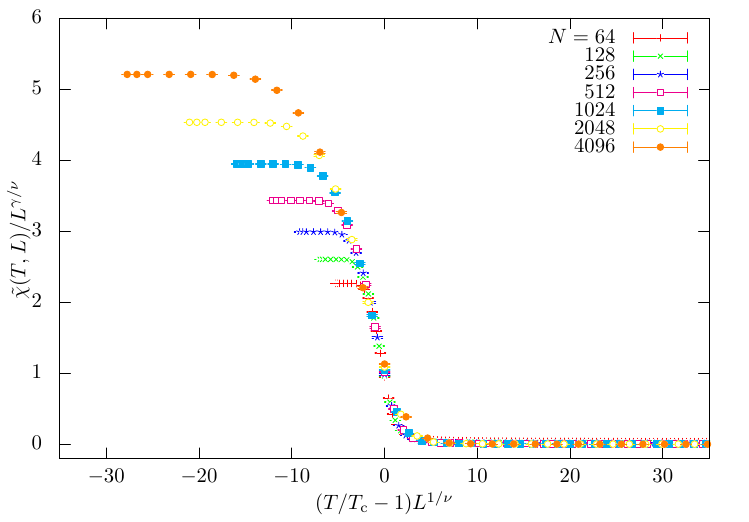}
	\caption{Finite-size scaling of $\tilde{\chi}$ with $T_{\mathrm{c}} = 0.24, \nu = 2.50$, and $\gamma = 2.00$. We set $K = 20, J = 1.0, q = 10^{-2.0}, t = 0.0$, and $\epsilon = 0.00$. We varied $N = 64, 128, \dots, 4096$.}
	\label{main_fig_finite_size_scaling_K=20_R=1_q_-2.0_e=0.00_001_001}
\end{figure}

Figure~\ref{main_fig_correlation_K=20_R=1_q_-2.0_e=0.00_001_001} shows the correlation function.
The plots for $k_{\mathrm{B}} T = 0.1$ and $0.2$ (i.e., $k_{\mathrm{B}} T < 0.24$) exhibit a nearly linear behavior, whereas those for $k_{\mathrm{B}} T > 0.24$ decay rapidly.
Figure~\ref{main_fig_finite_size_scaling_K=20_R=1_q_-2.0_e=0.00_001_001} shows the result of the finite-size scaling, in which the critical exponents are determined via the grid-search method.

\begin{figure}[t]
	\centering
	\includegraphics[scale=0.60]{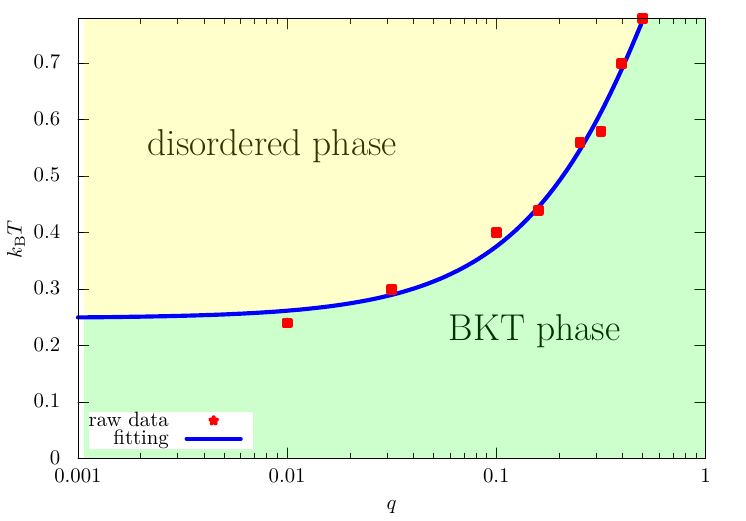}
	\caption{Phase diagram for $K = 20, J = 1.0, t = 0.0$, and $\epsilon = 0.00$. The horizontal and vertical axes represent the parameter $q$, which controls the probability of selecting production rules, and the critical temperature $k_{\mathrm{B}} T$, respectively.}
	\label{main_fig_q_phase_diagram_K=20_R=1_e=0.00_001_001}
\end{figure}

\begin{figure}[t]
	\centering
	\includegraphics[scale=0.60]{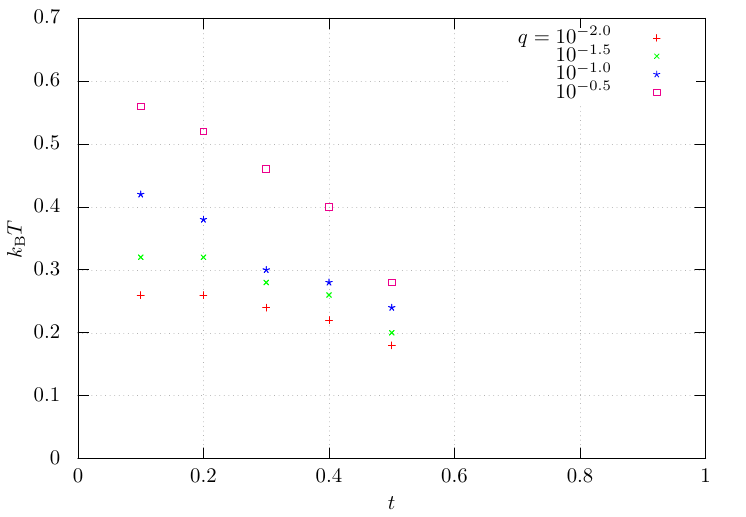}
	\caption{Phase diagram for $K = 20, J = 1.0$, and $\epsilon = 0.00$. The horizontal and vertical axes represent the parameter $t$, which controls the probability of selecting production rules, and the critical temperature $k_{\mathrm{B}} T$, respectively.}
	\label{main_fig_t_phase_diagram_K=20_R=1_e=0.00_001_001}
\end{figure}

In Figs.~\ref{main_fig_mag_sus_Bin_K=20_R=1_q_-2.0_e=0.00_001_001}, \ref{main_fig_correlation_K=20_R=1_q_-2.0_e=0.00_001_001} and \ref{main_fig_finite_size_scaling_K=20_R=1_q_-2.0_e=0.00_001_001}, we fixed $q = 10^{-2.0}$ and $t = 0.0$; to see the $q$-, $t$-dependence of the model, Eq.~\eqref{main_eq_rule_003_001}, we show phase diagrams in Figs.~\ref{main_fig_q_phase_diagram_K=20_R=1_e=0.00_001_001} and \ref{main_fig_t_phase_diagram_K=20_R=1_e=0.00_001_001} estimating the critical temperatures for various values of $q$ and $t$.
According to the forms of the Binder parameter in Fig.~\ref{main_fig_mag_sus_Bin_K=20_R=1_q_-2.0_e=0.00_001_001}(lower) and the correlation function in Fig.~\ref{main_fig_correlation_K=20_R=1_q_-2.0_e=0.00_001_001}, we insist that BKT phase is realized, as shown in Fig.~\ref{main_fig_q_phase_diagram_K=20_R=1_e=0.00_001_001}~\cite{Toji_001}.
Furthermore, we see an interesting phenomenon in Fig.~\ref{main_fig_t_phase_diagram_K=20_R=1_e=0.00_001_001}, that is, the sudden drops of the critical temperature at $t = 0.5$.
These points will be elaborated more in the following section.

\section{Discussions} \label{main_sec_discussion_001_001}

All three panels in Fig.~\ref{main_fig_mag_sus_Bin_K=20_R=1_q_-2.0_e=0.00_001_001} show the existence of a phase transition.
Although the singularity of the magnetization and the divergence of the susceptibility are weak due to finite-size effects, we can see that these tendencies become stronger as the system size $N$ increases.
The correlation function is shown in Fig.~\ref{main_fig_correlation_K=20_R=1_q_-2.0_e=0.00_001_001}.
In conventional second-order phase transitions, it is known that the correlation function decays polynomially with distance at the transition temperature.
However, in this figure, the correlation function decays polynomially below the transition temperature $k_{\mathrm{B}} T_* = 0.24$.
Thus, critical phenomena occur not only at the critical temperature but also below it.
This characteristic is identical to that in the BKT transition.

The BKT transition is known as a phenomenon that occurs in the $XY$ model, but it is also known to occur in one-dimensional models with long-range interactions~\cite{kosterlitz2016kosterlitz}.
However, our model has only short-range interactions.
We consider that the key to resolving this controversy is the process of increasing spin variables, which is unique to language models.
Even if the two spins do not interact when they are far apart in the large system, they may have interacted when the system size was still small, that is, at the beginning of the sentence generation, and this history may remain.
This may realize effective long-range interactions.

Finally, Fig.~\ref{main_fig_t_phase_diagram_K=20_R=1_e=0.00_001_001} shows the $k_{\mathrm{B}} T$--$t$ phase diagram.
We show it for various values of $q$.
For all $q$, phase transitions occur at finite temperature below $t=0.5$, but no phase transitions occur for $t > 0.5$.
This can be understood by recalling how the production rules are chosen.
For the $t < 0.5$ regime, the rule $X \to YZ$ is applied more frequently than $X \to XX$, so the increasing rate of non-terminal symbols tends to be faster than the speed of decrease of non-terminal symbols.
Therefore, each sentence tends to be long and the model behaves like $t=0$.
In contrast, for $t > 0.5$ regime, the decrease dominates over the increase, so short sentences are generated frequently.
In our simulations, we set the start symbol of each sentence after the first one to be the last symbol of the previous sentence to introduce correlations among sentences.
With this correlation, the order of whole sentences is lost, and phase transitions do not occur.
It is nontrivial whether phase transition occurs at $t=0.5$, but we observe phase transition at $t=0.5$.
Moreover, it is also nontrivial that the critical temperatures decrease as $t$ increases.

\section{Conclusions} \label{main_sec_conclusion_001_001}

We constructed the CSG model inspired by the one-dimensional short-range Potts model and observed a phase transition in the language generated by this model.
Moreover, we confirmed that this transition is consistent with a BKT transition based on the behavior of the Binder parameter and the correlation functions.
Since it is widely believed that one-dimensional equilibrium systems with short-range interactions do not exhibit phase transitions, the linguistic process investigated in this study appears to induce nonequilibrium effects or effective long-range interactions.
We believe that this work opens a possible avenue toward studying language from the perspective of nonequilibrium statistical physics.

\begin{acknowledgments}
	H.M. was supported by JSPS KAKENHI Grant Numbers JP25H01499, JP26H01783, and JP26K17043.
\end{acknowledgments}

\appendix
\renewcommand{\theequation}{\Alph{section}.\arabic{equation}}

\section{Numerical simulations} \label{appendix_sec_numerical_simulations_001_001}

We provide additional numerical simulations to support the claims in the main text.

\subsection{Case of $K = 2, \epsilon = 0.00$ and $t = 0.0$}

\subsubsection{$k_{\mathrm{B}} T$--$q$ phase diagram}

\begin{figure}[t]
	\centering
	\includegraphics[scale=0.54]{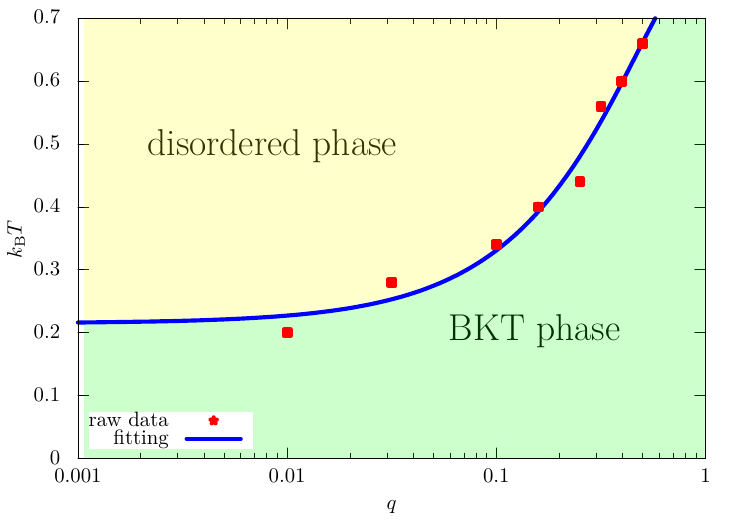}
	\caption{Phase diagram for $K = 2, J = 1.0, t = 0.0$, and $\epsilon = 0.00$. The horizontal and vertical axes represent the parameter $q$, which controls the probability of selecting production rules, and the critical temperature $k_{\mathrm{B}} T$, respectively.}
	\label{appendix_fig_phase_diagram_K=2_R=1_e=0.00_001_001}
\end{figure}

In Fig.~\ref{appendix_fig_phase_diagram_K=2_R=1_e=0.00_001_001},
we show a $k_{\mathrm{B}} T$--$q$ phase diagram of $K = 2$ and $\epsilon = 0.00$.

\subsubsection{Magnetization, susceptibility and Binder parameter}

\begin{figure}[t]
	\centering
	\includegraphics[scale=0.54]{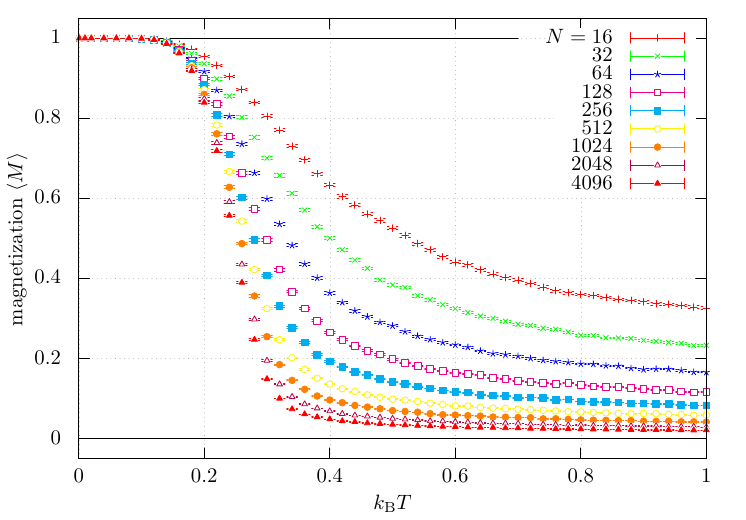}
	\includegraphics[scale=0.54]{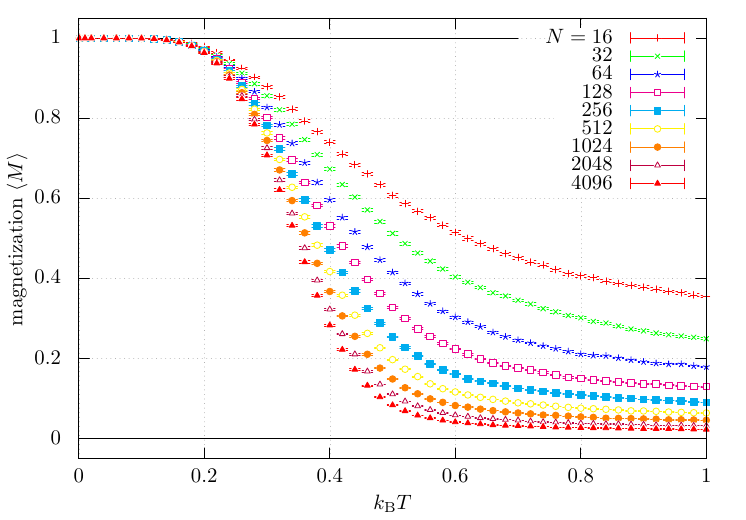}
	\caption{Temperature dependence of the magnetization, Eq.~\eqref{main_eq_magnetization_001_001}. The top panel is the result for $q = 10^{-2.0}$, and the bottom panel is the result for $q = 10^{-1.0}$. We set $K = 2, J = 1.0, t = 0.0$, and $\epsilon = 0.00$. We show the results for various system sizes $N = 16, 32, \dots, 4096$, and the curves for different $N$ are overlaid in each panel.}
	\label{appendix_fig_mag_K=2_R=1_e=0.00_001_001}
\end{figure}

\begin{figure}[t]
	\centering
	\includegraphics[scale=0.54]{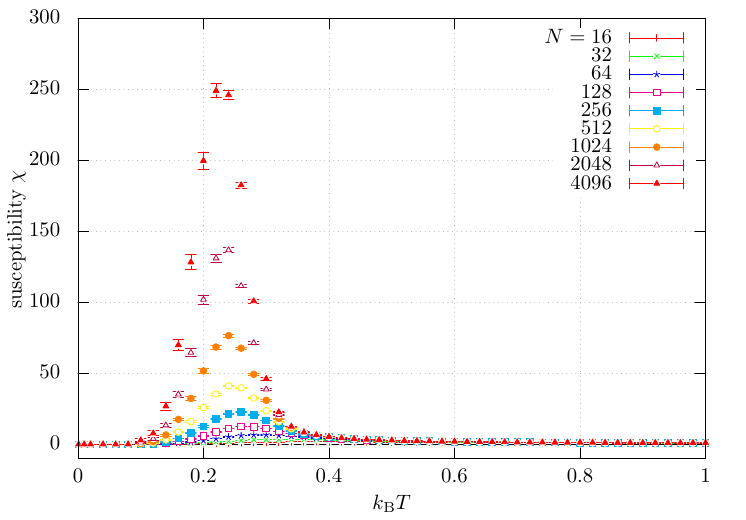}
	\includegraphics[scale=0.54]{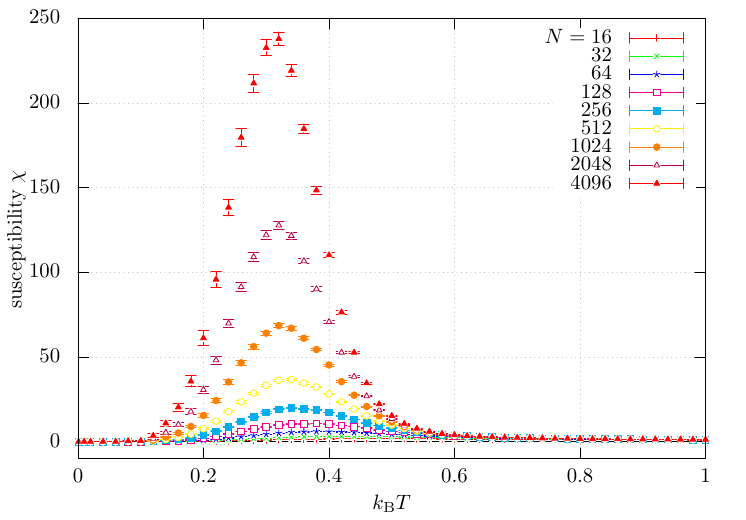}
	\caption{Temperature dependence of the susceptibility, Eq.~\eqref{main_eq_susceptibility_001_001}. The top panel is the result for $q = 10^{-2.0}$, and the bottom panel is the result for $q = 10^{-1.0}$. We set $K = 2, J = 1.0, t = 0.0$, and $\epsilon = 0.00$. We show the results for various system sizes $N = 16, 32, \dots, 4096$, and the curves for different $N$ are overlaid in each panel.}
	\label{appendix_fig_sus_K=2_R=1_e=0.00_001_001}
\end{figure}

\begin{figure}[t]
	\centering
	\includegraphics[scale=0.54]{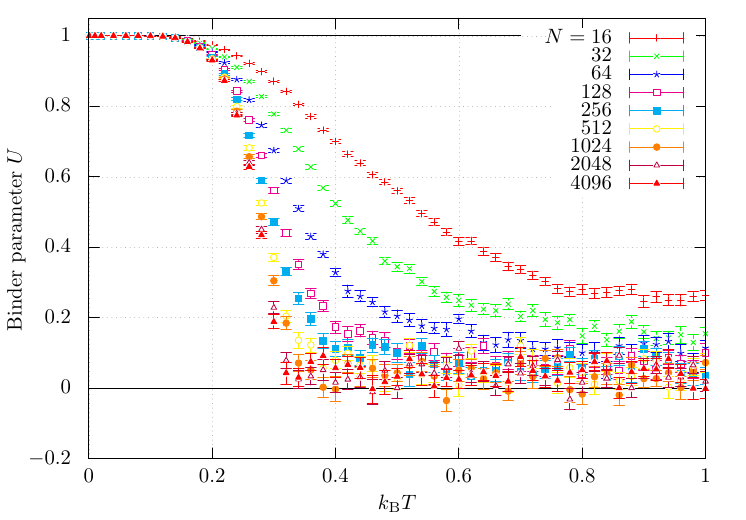}
	\includegraphics[scale=0.54]{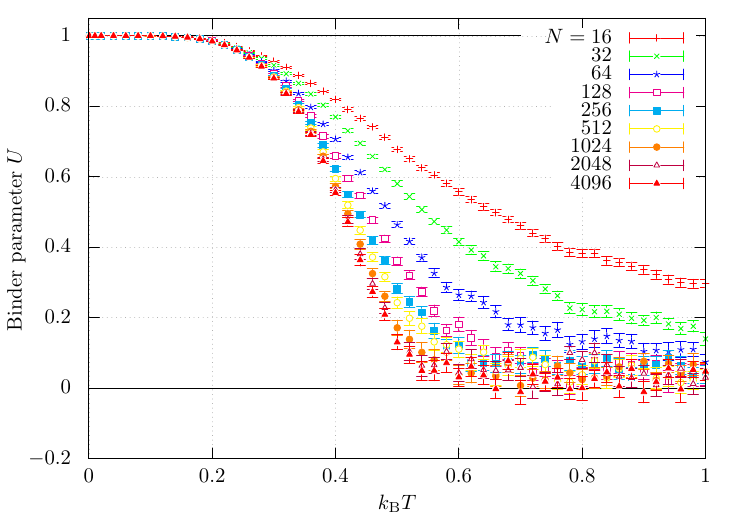}
	\caption{Temperature dependence of the Binder parameter, Eq.~\eqref{main_eq_Binder_parameter_001_001}. The top panel is the result for $q = 10^{-2.0}$, and the bottom panel is the result for $q = 10^{-1.0}$. We set $K = 2, J = 1.0, t = 0.0$, and $\epsilon = 0.00$. We show the results for various system sizes $N = 16, 32, \dots, 4096$, and the curves for different $N$ are overlaid in each panel.}
	\label{appendix_fig_Binder_K=2_R=1_e=0.00_001_001}
\end{figure}

Below, we compare the results for $q = 10^{-2.0}$ and $q = 10^{-1.0}$.
The transition temperatures are estimated to be $k_{\mathrm{B}} T = 0.20$ for $q = 10^{-2.0}$
and $k_{\mathrm{B}} T = 0.34$ for $q = 10^{-1.0}$.
The magnetization is shown in Fig.~\ref{appendix_fig_mag_K=2_R=1_e=0.00_001_001},
the susceptibility is shown in Fig.~\ref{appendix_fig_sus_K=2_R=1_e=0.00_001_001},
and the Binder parameter is shown in Fig.~\ref{appendix_fig_Binder_K=2_R=1_e=0.00_001_001}.


We also show the system-size dependence of the magnetization, susceptibility, and Binder parameter.
Figure~\ref{appendix_fig_mag_K=2_R=1_e=0.00_001_001} shows the system-size dependence of the magnetization, Fig.~\ref{appendix_fig_sus_K=2_R=1_e=0.00_001_001} shows that of the susceptibility, and Fig.~\ref{appendix_fig_Binder_K=2_R=1_e=0.00_001_001} shows that of the Binder parameter.
\begin{figure}[t]
	\centering
	\includegraphics[scale=0.54]{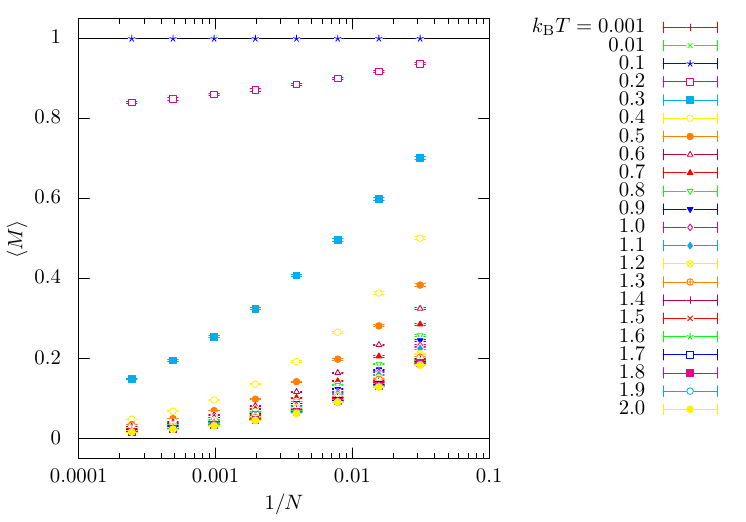}
	\includegraphics[scale=0.54]{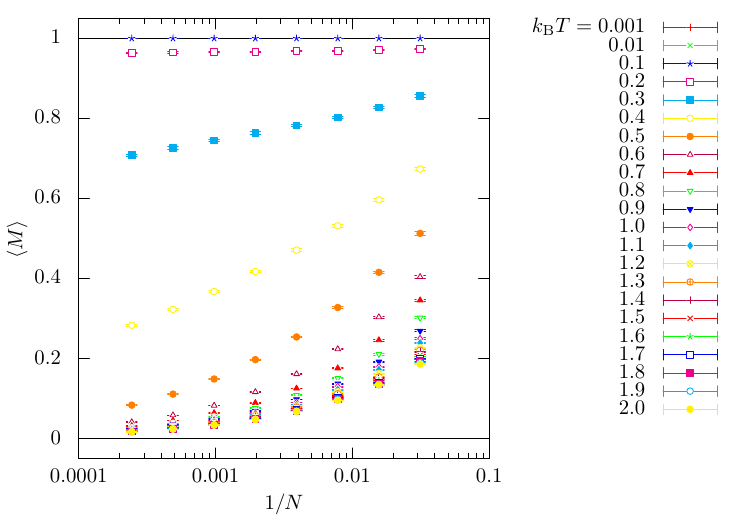}
	\caption{The system-size dependence of the magnetization, Eq.~\eqref{main_eq_magnetization_001_001}. The top panel is the result for $q = 10^{-2.0}$, and the bottom panel is the result for $q = 10^{-1.0}$. We set $K = 2, J = 1.0, t = 0.0$, and $\epsilon = 0.00$. We show the results for various $k_{\mathrm{B}} T = 0.001, 0.01, 0.1, 0.2, \dots, 2.0$, and the curves for different $k_{\mathrm{B}} T$ are overlaid.}
	\label{appendix_fig_size_dependence_mag_K=2_R=1_e=0.00_001_001}
\end{figure}
\begin{figure}[t]
	\centering
	\includegraphics[scale=0.54]{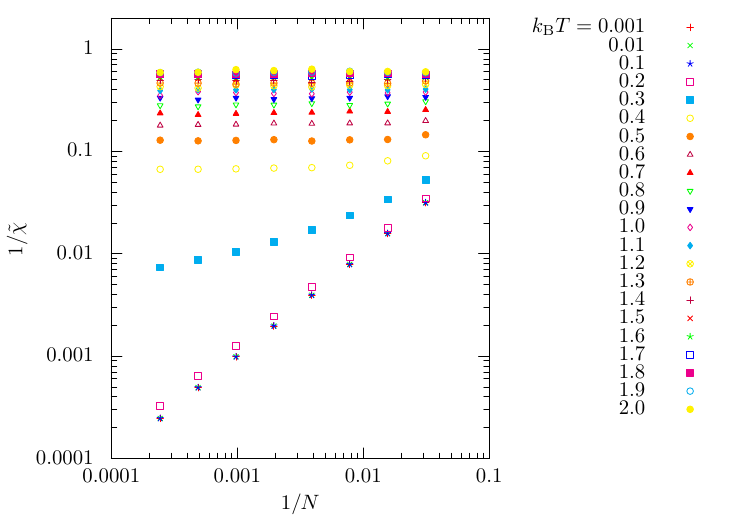}
	\includegraphics[scale=0.54]{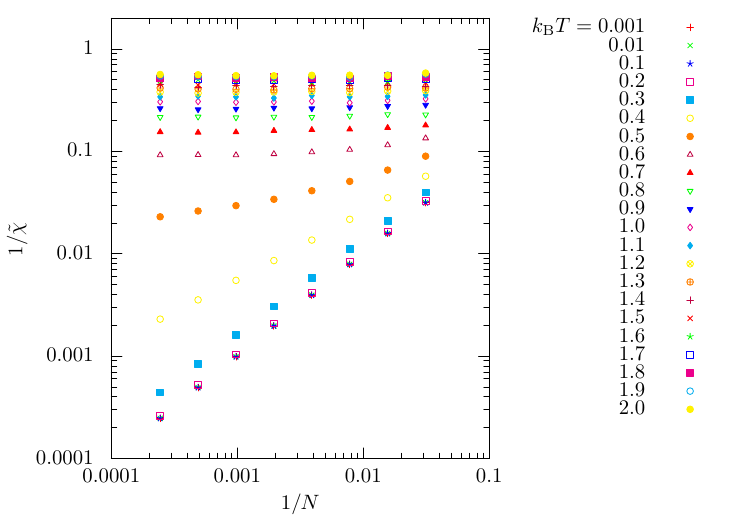}
	\caption{The system-size dependence of the susceptibility, Eq.~\eqref{main_eq_susceptibility_001_001}. The top panel is the result for $q = 10^{-2.0}$, and the bottom panel is the result for $q = 10^{-1.0}$. We set $K = 2, J = 1.0, t = 0.0$, and $\epsilon = 0.00$. We show the results for various $k_{\mathrm{B}} T = 0.001, 0.01, 0.1, 0.2, \dots, 2.0$, and the curves for different $k_{\mathrm{B}} T$ are overlaid.}
	\label{appendix_fig_size_dependence_sus_K=2_R=1_e=0.00_001_001}
\end{figure}
\begin{figure}[t]
	\centering
	\includegraphics[scale=0.54]{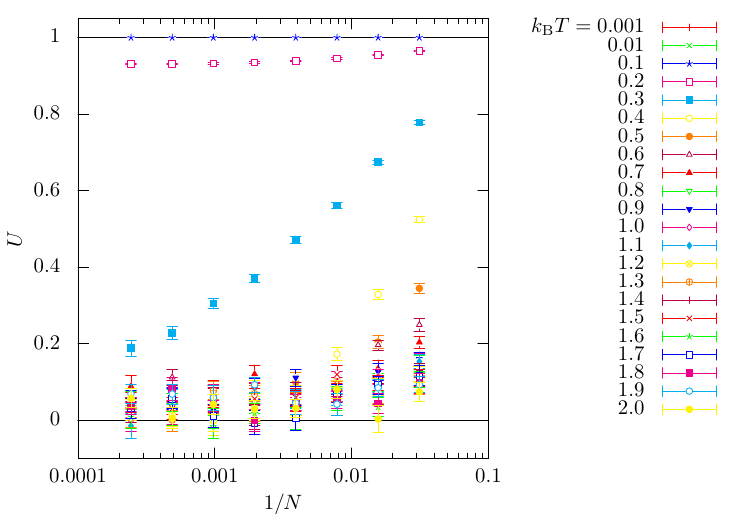}
	\includegraphics[scale=0.54]{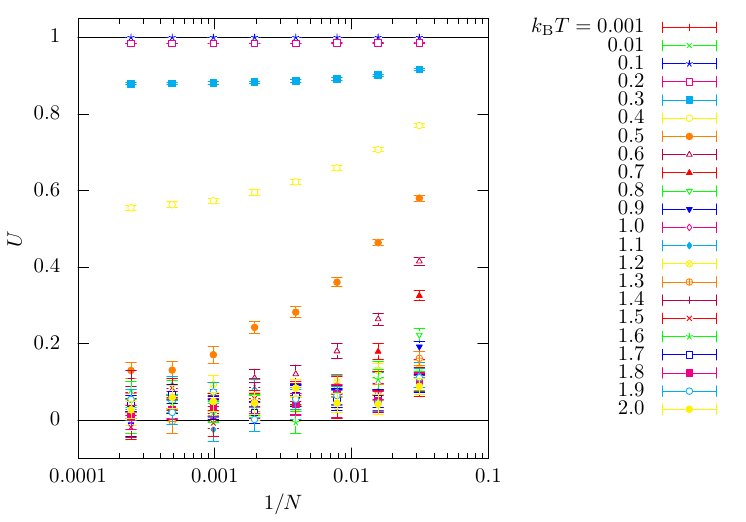}
	\caption{The system-size dependence of the Binder parameter, Eq.~\eqref{main_eq_Binder_parameter_001_001}. The top panel is the result for $q = 10^{-2.0}$, and the bottom panel is the result for $q = 10^{-1.0}$. We set $K = 2, J = 1.0, t = 0.0$, and $\epsilon = 0.00$. We show the results for various $k_{\mathrm{B}} T = 0.001, 0.01, 0.1, 0.2, \dots, 2.0$, and the curves for different $k_{\mathrm{B}} T$ are overlaid.}
	\label{appendix_fig_size_dependence_Binder_K=2_R=1_e=0.00_001_001}
\end{figure}

\subsubsection{Histogram of the magnetization} \label{appendix_subsec_histogram_K=2_R=1_e=0.00_001}

\begin{figure}[t]
	\centering
	\includegraphics[scale=0.54]{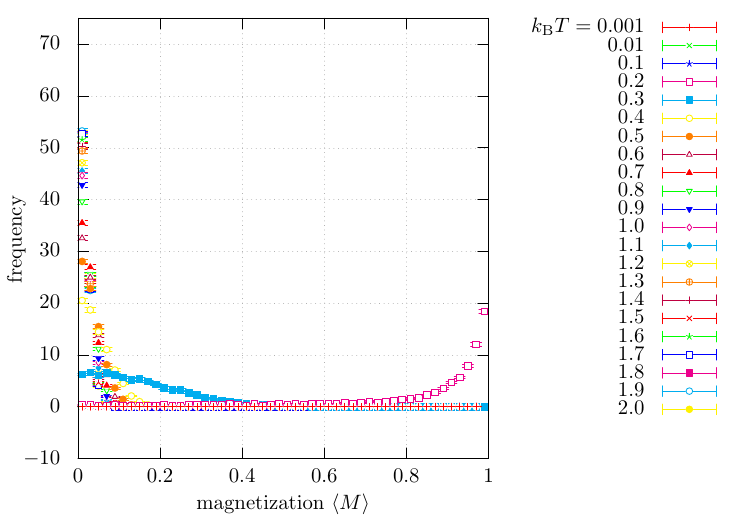}
	\includegraphics[scale=0.54]{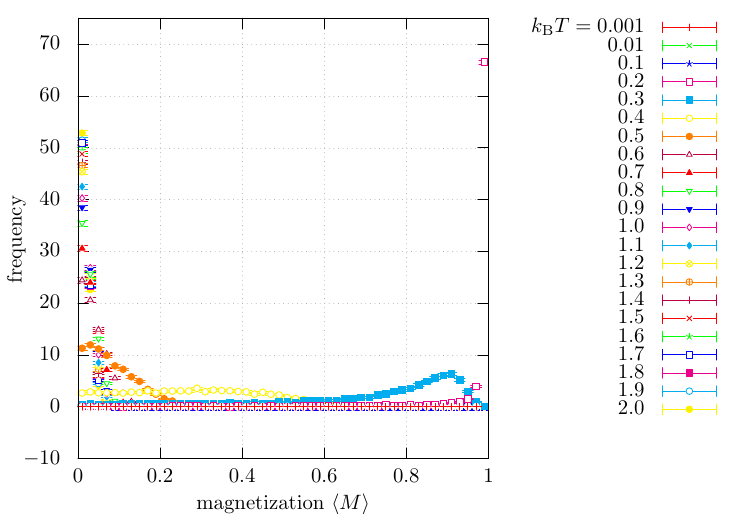}
	\caption{The histogram of the magnetization. The top panel is the result for $q = 10^{-2.0}$, and the bottom panel is the result for $q = 10^{-1.0}$. We set $K = 2$, $J = 1.0$, $t = 0.0$, and $\epsilon = 0.00$. We show the results for various $k_{\mathrm{B}} T = 0.001, 0.01, 0.1, 0.2, \dots, 2.0$, and the curves for different $k_{\mathrm{B}} T$ are overlaid.}
	\label{appendix_fig_histogram_K=2_R=1_e=0.00_001_001}
\end{figure}

The histogram of the magnetization is shown in Fig.~\ref{appendix_fig_histogram_K=2_R=1_e=0.00_001_001}.

\subsubsection{Typical configurations of symbols}

\begin{figure}[t]
	\centering
	\includegraphics[scale=0.54]{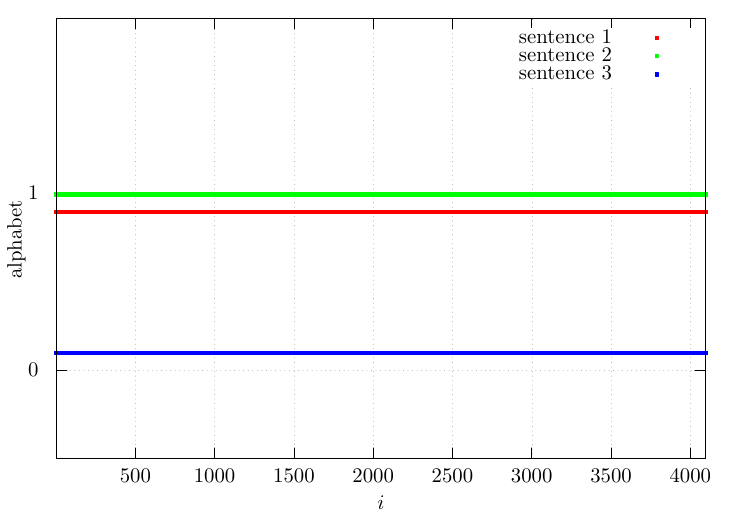}
	\includegraphics[scale=0.54]{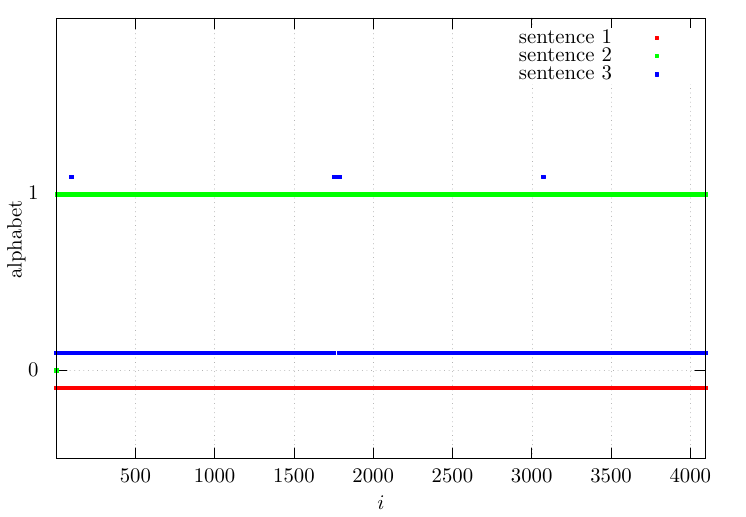}
	\includegraphics[scale=0.54]{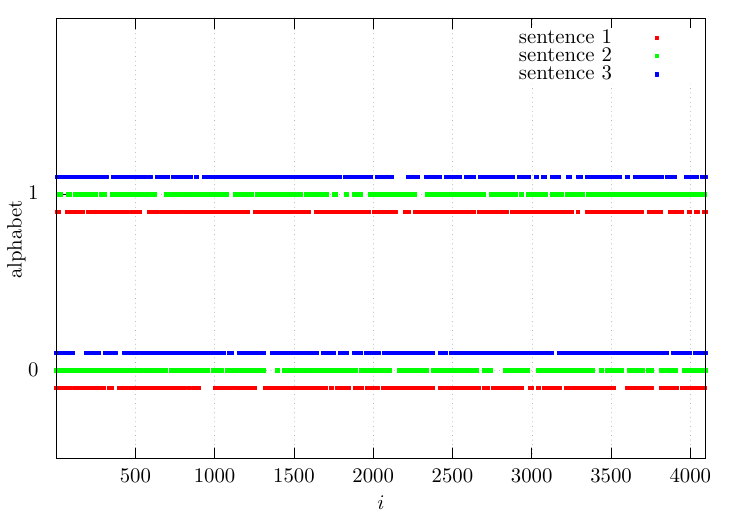}
	\caption{Typical configurations of symbols. The upper panel shows an example sentence at a temperature below the transition ($k_{\mathrm{B}} T = 0.001$) with magnetization $0.98 \leq M < 1.00$. The middle panel shows an example sentence at a temperature just above the transition ($k_{\mathrm{B}} T = 0.20$) with magnetization $0.98 \leq M < 1.00$. The lower panel shows an example sentence at a temperature above the transition ($k_{\mathrm{B}} T = 0.40$) with magnetization $0.00 \leq M < 0.02$. We set $K = 2$, $J = 1.0$, $q=10^{-2.0}$, $t = 0.0$, and $\epsilon = 0.00$.}
	\label{appendix_fig_configuration_K=2_q=10_-2.0_R=1_e=0.00_001_001}
\end{figure}

\begin{figure}[t]
	\centering
	\includegraphics[scale=0.54]{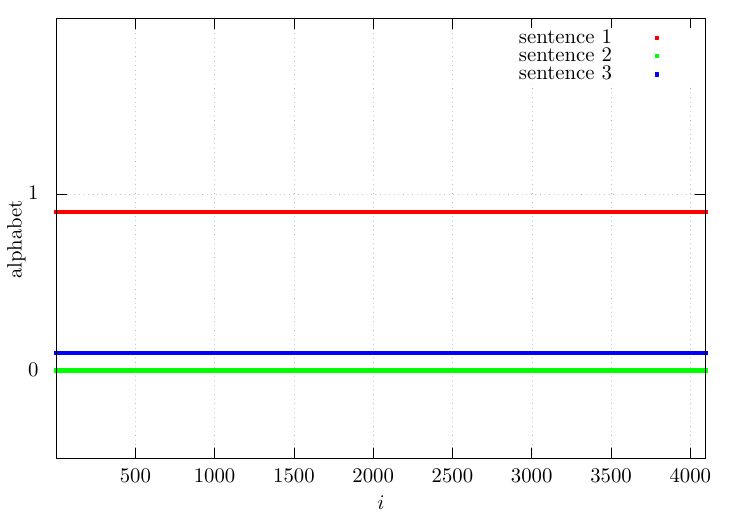}
	\includegraphics[scale=0.54]{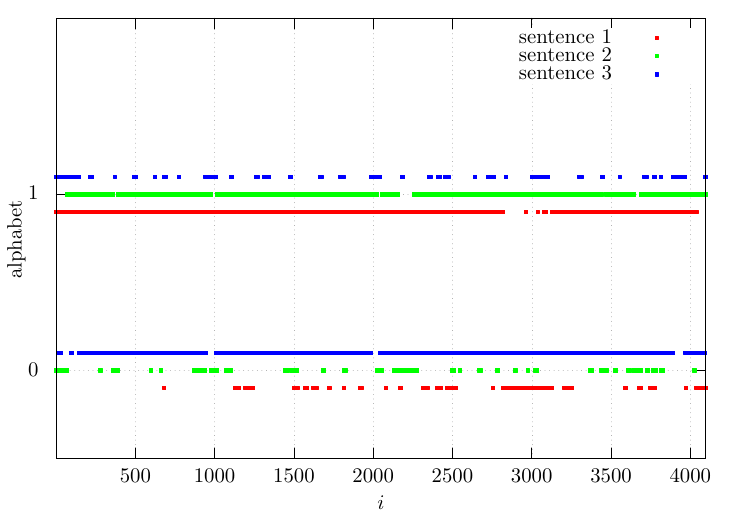}
	\includegraphics[scale=0.54]{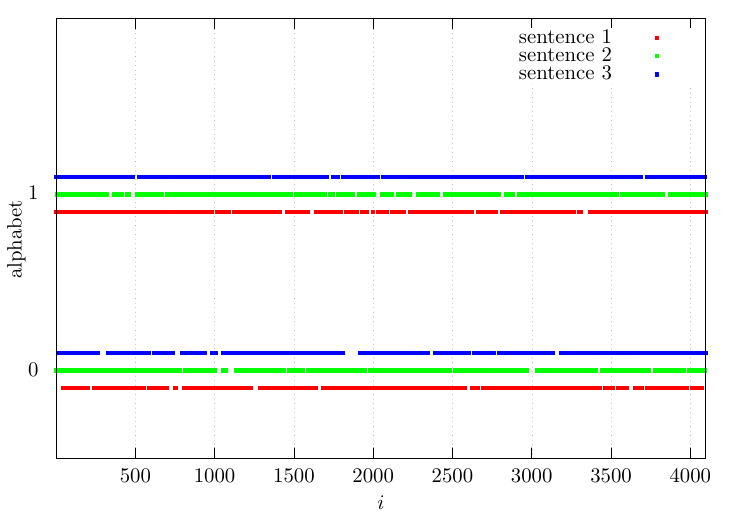}
	\caption{Typical configurations of symbols. The upper panel shows an example sentence at a temperature below the transition ($k_{\mathrm{B}} T = 0.14$) with magnetization $0.98 \leq M < 1.00$. The middle panel shows an example sentence at a temperature just above the transition ($k_{\mathrm{B}} T = 0.34$) with magnetization $0.72 \leq M < 0.74$. The lower panel shows an example sentence at a temperature above the transition ($k_{\mathrm{B}} T = 0.54$) with magnetization $0.00 \leq M < 0.02$. We set $K = 2$, $J = 1.0$, $q=10^{-1.0}$, $t = 0.0$, and $\epsilon = 0.00$.}
	\label{appendix_fig_configuration_K=2_q=10_-1.0_R=1_e=0.00_001_001}
\end{figure}

By examining the histograms shown in Sec.~\ref{appendix_subsec_histogram_K=2_R=1_e=0.00_001}, we can identify, for each temperature, the most frequent value of the magnetization.
We regard this value as the typical magnetization at that temperature, and we show, as examples, the configurations of the generated text whose magnetization takes this typical value in Figs.~\ref{appendix_fig_configuration_K=2_q=10_-2.0_R=1_e=0.00_001_001} and \ref{appendix_fig_configuration_K=2_q=10_-1.0_R=1_e=0.00_001_001}.
The temperatures shown correspond to values below the transition temperature, just above the transition temperature, and above the transition temperature.

\subsubsection{Correlation function}

\begin{figure}[t]
	\centering
	\includegraphics[scale=0.54]{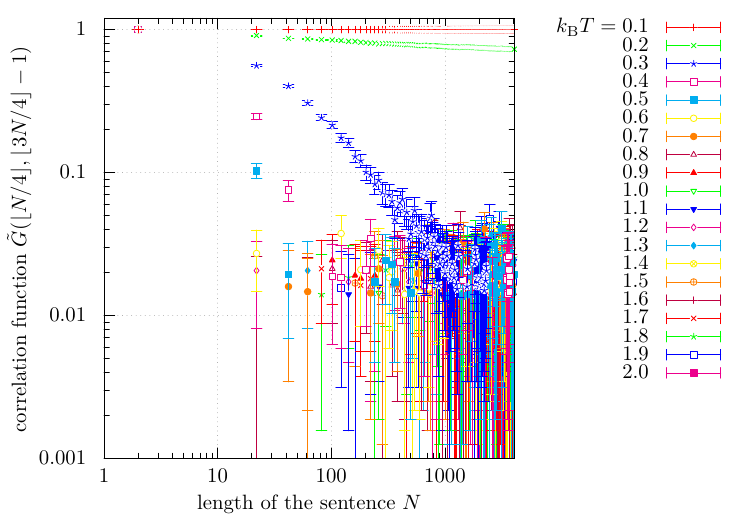}
	\includegraphics[scale=0.54]{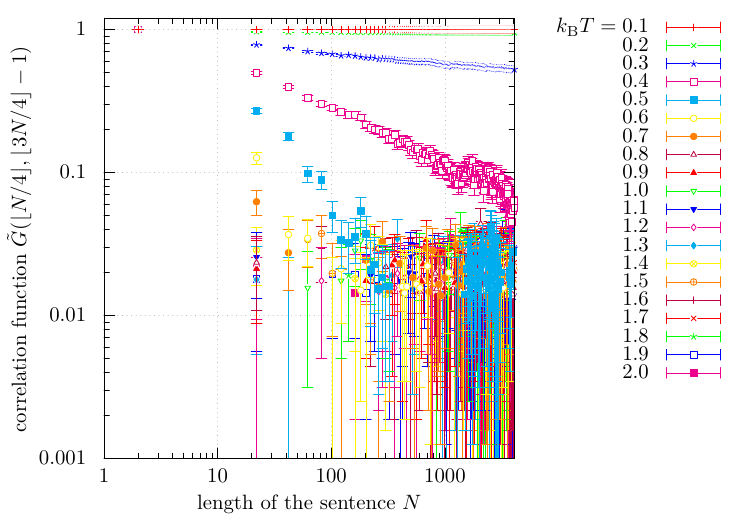}
	\caption{The correlation function~, Eq.~\eqref{main_eq_correlation_function_001_002}, with $i = \lfloor N/4 \rfloor$ and $j = \lfloor 3N/4 \rfloor - 1$. The top panel is the result for $q = 10^{-2.0}$, and the bottom panel is the result for $q = 10^{-1.0}$. We set $K = 2, J = 1.0, t = 0.0$, and $\epsilon = 0.00$. We show the results for various $k_{\mathrm{B}} T = 0.1, 0.2, \dots, 2.0$, and the curves for different $k_{\mathrm{B}} T$ are overlaid.}
	\label{appendix_fig_correlation_K=2_R=1_e=0.00_001_001}
\end{figure}

The data of the correlation function $\tilde{G}(\lfloor N/4 \rfloor, \lfloor 3N/4 \rfloor - 1)$ are shown in Fig.~\ref{appendix_fig_correlation_K=2_R=1_e=0.00_001_001}.

\subsubsection{Mutual information}

\begin{figure}[t]
	\centering
	\includegraphics[scale=0.54]{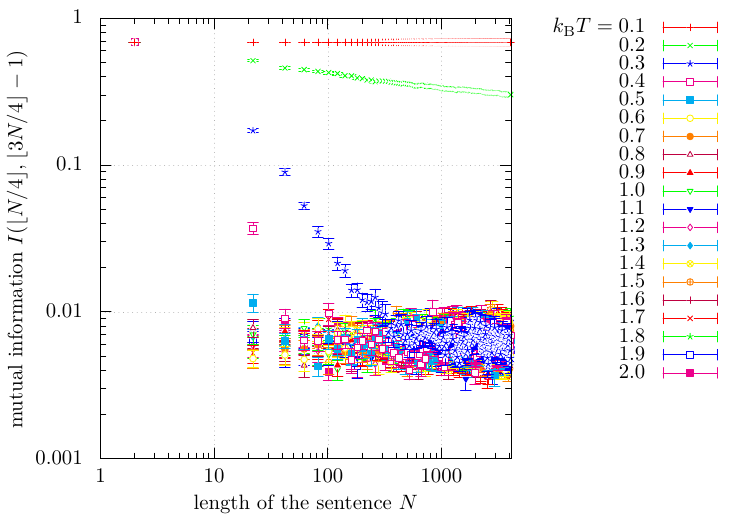}
	\includegraphics[scale=0.54]{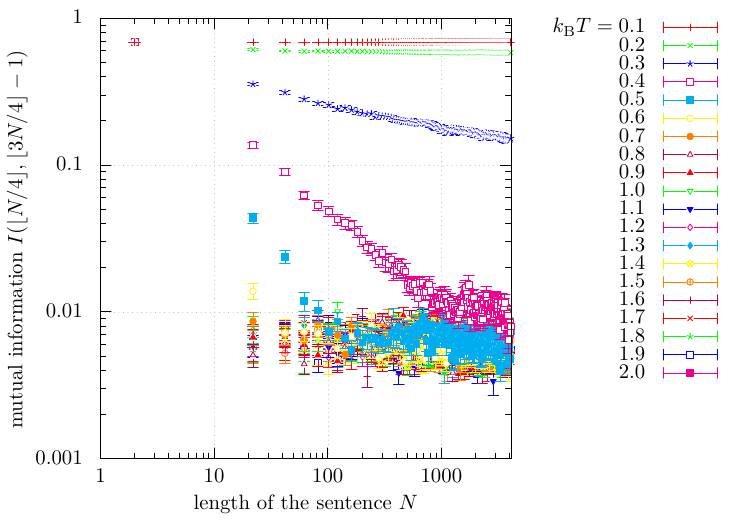}
	\caption{The mutual information $I(\lfloor N/4 \rfloor, \lfloor 3N/4 \rfloor - 1)$. The top panel is the result for $q = 10^{-2.0}$, and the bottom panel is the result for $q = 10^{-1.0}$. We set $K = 2, J = 1.0, t = 0.0$, and $\epsilon = 0.00$. We show the results for various $k_{\mathrm{B}} T = 0.1, 0.2, \dots, 2.0$, and the curves for different $k_{\mathrm{B}} T$ are overlaid.}
	\label{appendix_fig_mutual_information_K=2_R=1_e=0.00_001_001}
\end{figure}

The data of the mutual information $I(\lfloor N/4 \rfloor, \lfloor 3N/4 \rfloor - 1)$
are shown in Fig.~\ref{appendix_fig_mutual_information_K=2_R=1_e=0.00_001_001}.

\subsubsection{Finite-size scaling}

\begin{figure}[t]
	\centering
	\includegraphics[scale=0.54]{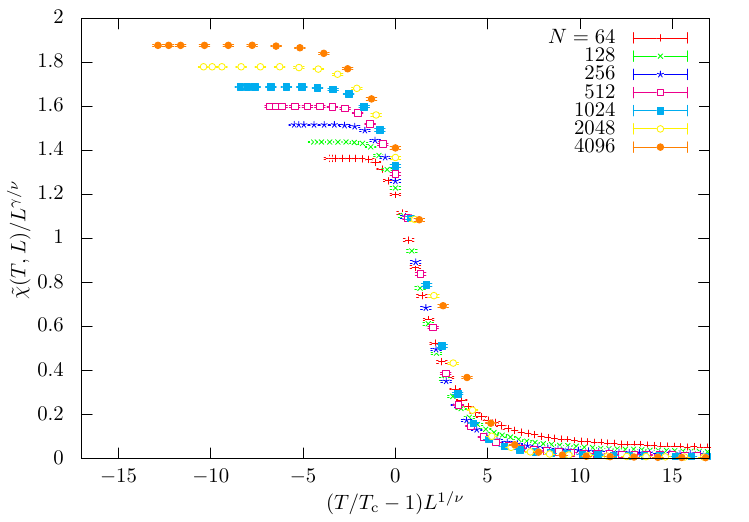}
	\includegraphics[scale=0.54]{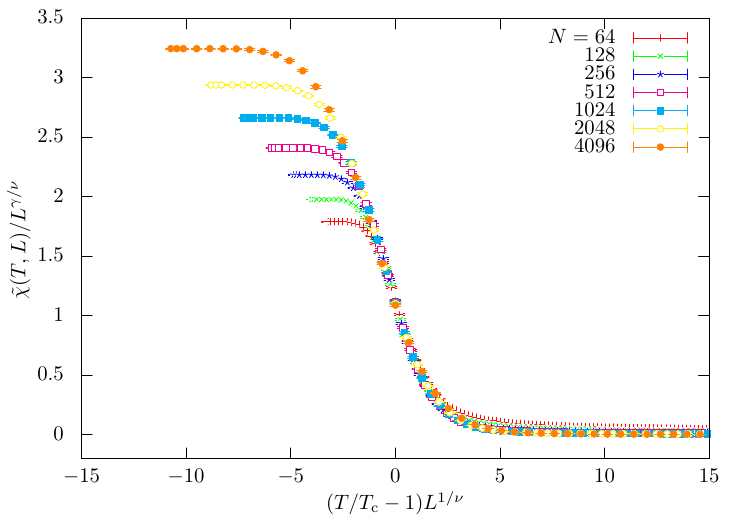}
	\caption{The finite-size scaling plot. The top panel is the result for $q = 10^{-2.0}$, and the bottom panel is the result for $q = 10^{-1.0}$. We plot $\tilde{\chi}(T,N) / N^{\gamma/\nu}$ versus $N^{1/\nu} t$. The critical exponents are set to $\nu = 3.25$ and $\gamma = 3.00$ for $q = 10^{-2.0}$, and to $\nu = 3.50$ and $\gamma = 3.00$ for $q = 10^{-1.0}$. We set $K = 2$, $J = 1.0$, $t = 0.0$, and $\epsilon = 0.00$. We show the results for $N = 64, 128, 256, 512, 1024, 2048, 4096$, and the curves for different $N$ are overlaid.}
	\label{appendix_fig_finite_size_scaling_K=2_R=1_e=0.00_001_001}
\end{figure}

\begin{figure}[t]
	\centering
	\includegraphics[scale=0.54]{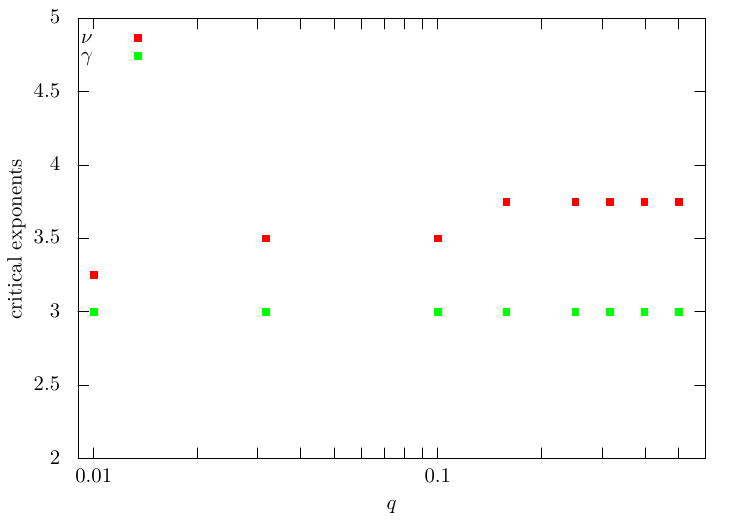}
	\caption{The $q$-dependence of the critical exponents $\nu$ and $\gamma$. We set $K = 2$, $J = 1.0$, $t = 0.0$, and $\epsilon = 0.00$. We show the results for $N = 64, 128, 256, 512, 1024, 2048, 4096$, and the curves for different $N$ are overlaid.}
	\label{appendix_fig_critical_exponents_q_K=2_R=1_e=0.00_001_001}
\end{figure}

The finite-size scaling plot is shown in Fig.~\ref{appendix_fig_finite_size_scaling_K=2_R=1_e=0.00_001_001}.
The $q$-dependence of the critical exponents $\gamma$ and $\nu$ is shown in Fig.~\ref{appendix_fig_critical_exponents_q_K=2_R=1_e=0.00_001_001}.

\clearpage

\subsection{Case of $K = 2, \epsilon = 0.00$, and $t \ne 0$}

\subsubsection{Magnetization, susceptibility and Binder parameter}

\begin{figure}[t]
	\centering
	\includegraphics[scale=0.54]{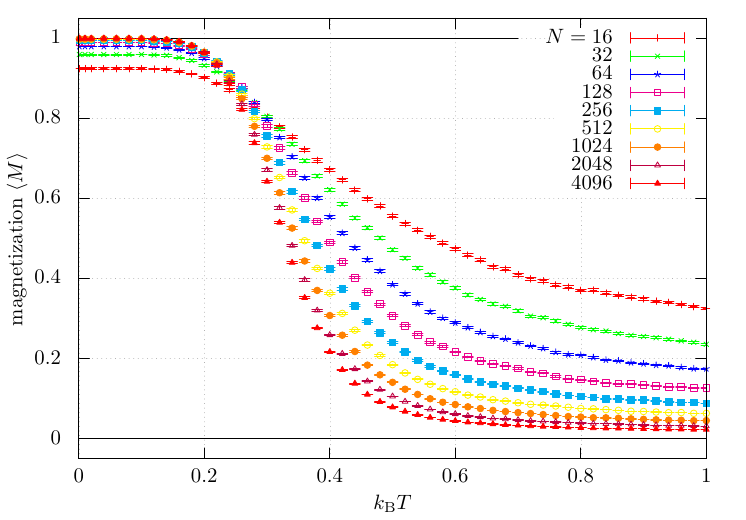}
	\includegraphics[scale=0.54]{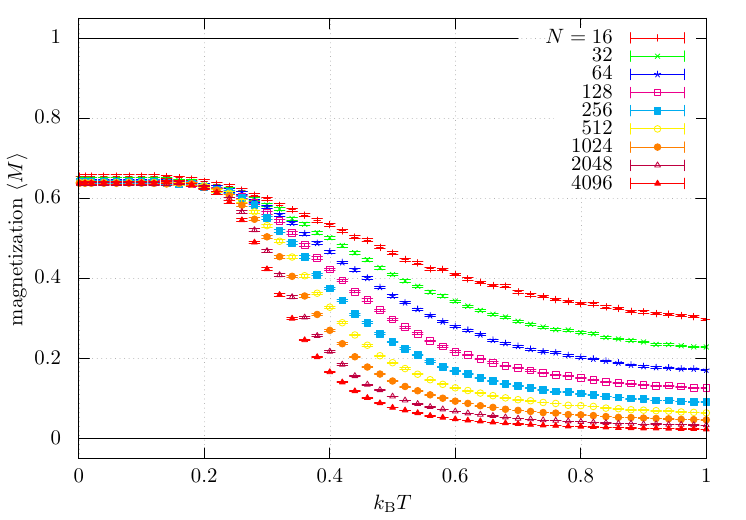}
	\includegraphics[scale=0.54]{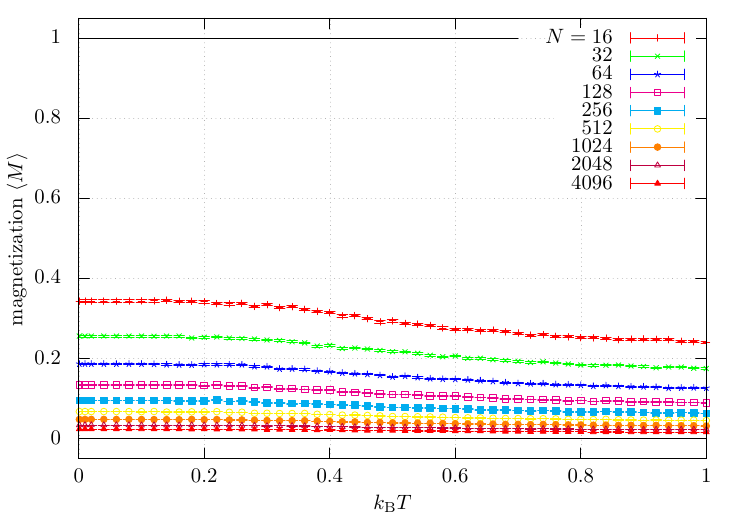}
	\caption{Temperature dependence of the magnetization, Eq.~\eqref{main_eq_magnetization_001_001}. The first panel is the result for $t = 0.3$, the second panel is the result for $t = 0.5$, and the third panel is the result for $t = 0.7$. We set $K = 2$, $J = 1.0$, $q = 10^{-2.0}$, and $\epsilon = 0.00$. We show the results for various system sizes $N = 16, 32, \dots, 4096$, and the curves for different $N$ are overlaid in each panel.}
	\label{appendix_fig_mag_K = 2_R=1_q_2.0_e=0.00_t!=0.0_001_001}
\end{figure}

\begin{figure}[t]
	\centering
	\includegraphics[scale=0.54]{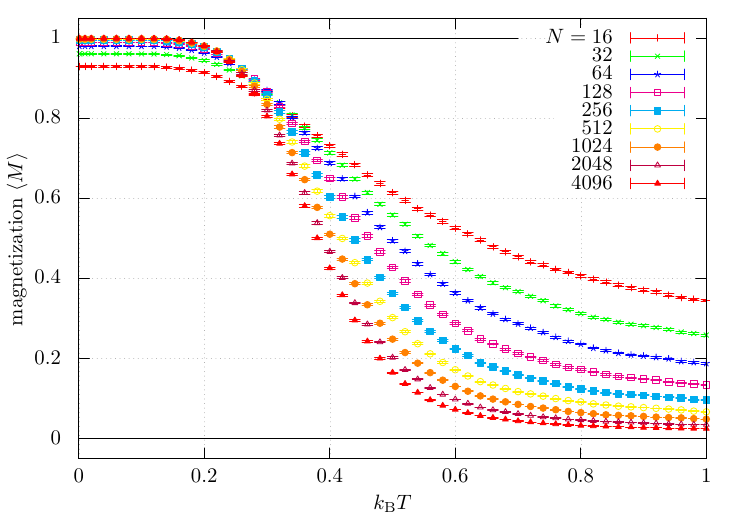}
	\includegraphics[scale=0.54]{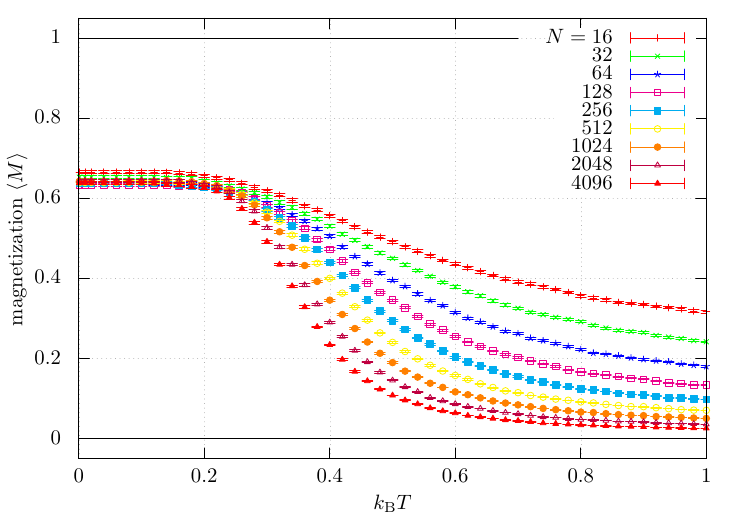}
	\includegraphics[scale=0.54]{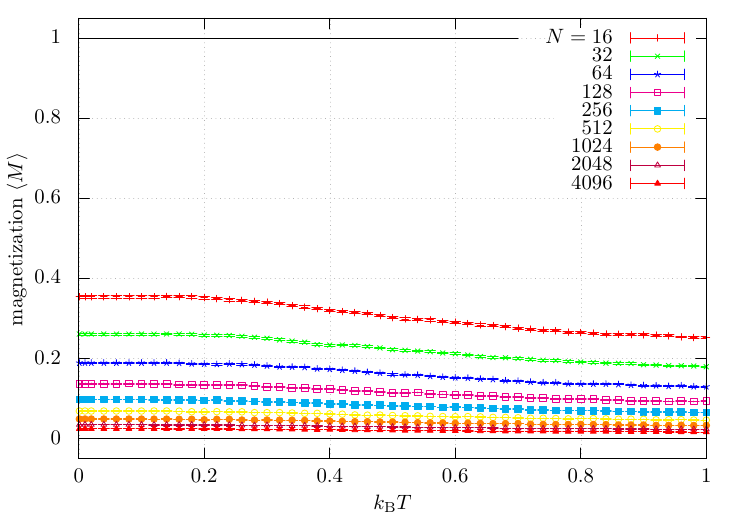}
	\caption{Temperature dependence of the magnetization, Eq.~\eqref{main_eq_magnetization_001_001}. The first panel is the result for $t = 0.3$, the second panel is the result for $t = 0.5$, and the third panel is the result for $t = 0.7$. We set $K = 2$, $J = 1.0$, $q = 10^{-1.0}$, and $\epsilon = 0.00$. We show the results for various system sizes $N = 16, 32, \dots, 4096$, and the curves for different $N$ are overlaid in each panel.}
	\label{appendix_fig_mag_K = 2_R=1_q_1.0_e=0.00_t!=0.0_001_001}
\end{figure}

\begin{figure}[t]
	\centering
	\includegraphics[scale=0.54]{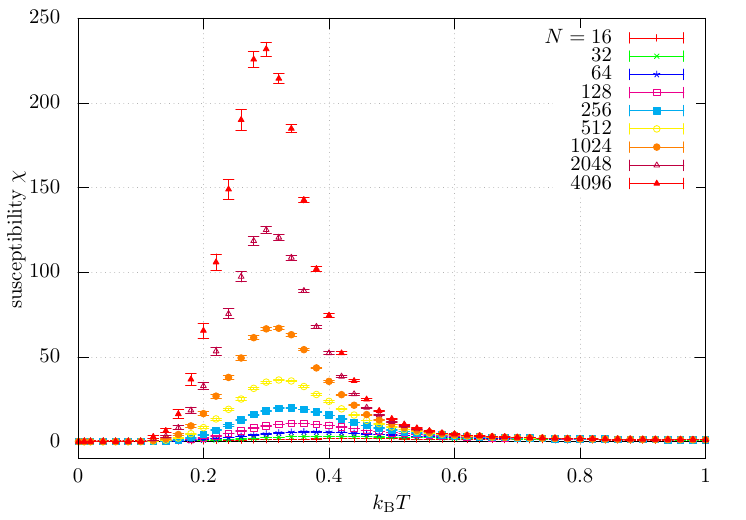}
	\includegraphics[scale=0.54]{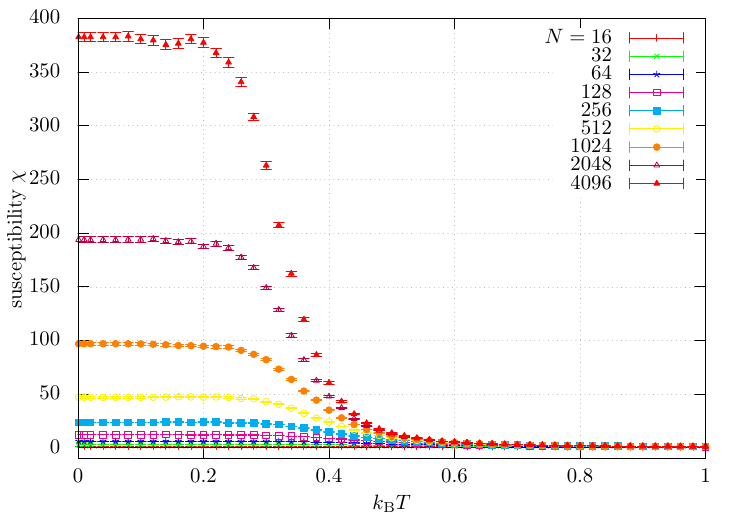}
	\includegraphics[scale=0.54]{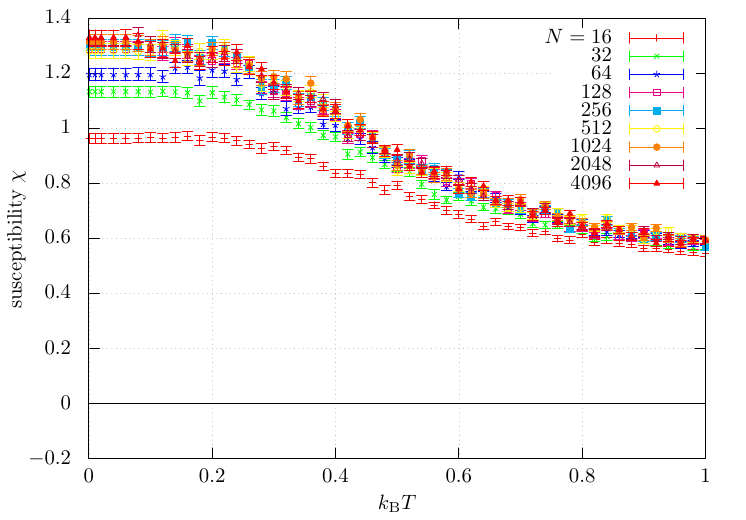}
	\caption{Temperature dependence of the susceptibility, Eq.~\eqref{main_eq_susceptibility_001_001}. The first panel is the result for $t = 0.3$, the second panel is the result for $t = 0.5$, and the third panel is the result for $t = 0.7$. We set $K = 2$, $J = 1.0$, $q = 10^{-2.0}$, and $\epsilon = 0.00$. We show the results for various system sizes $N = 16, 32, \dots, 4096$, and the curves for different $N$ are overlaid in each panel.}
	\label{appendix_fig_sus_K = 2_R=1_q_2.0_e=0.00_t!=0.0_001_001}
\end{figure}

\begin{figure}[t]
	\centering
	\includegraphics[scale=0.54]{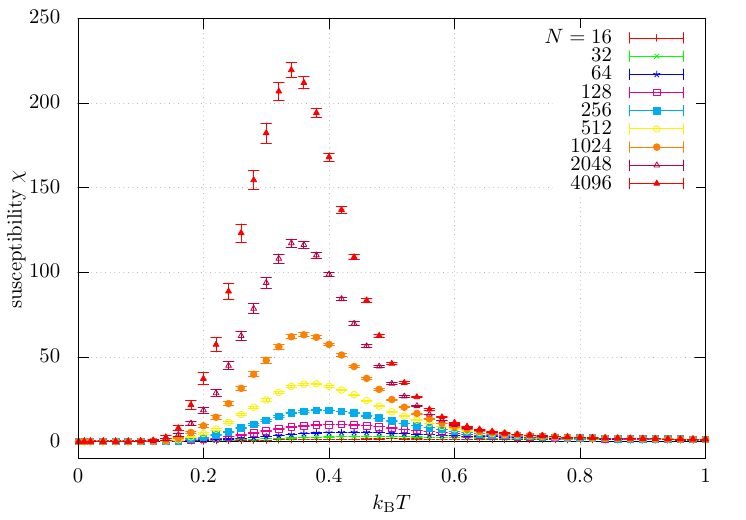}
	\includegraphics[scale=0.54]{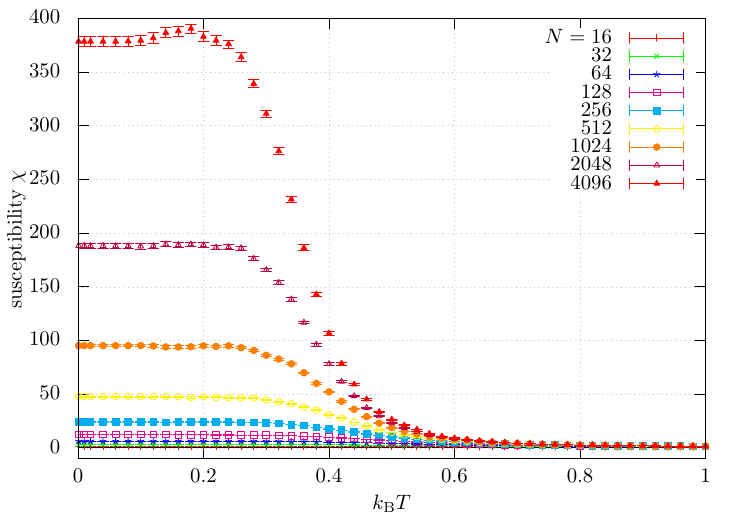}
	\includegraphics[scale=0.54]{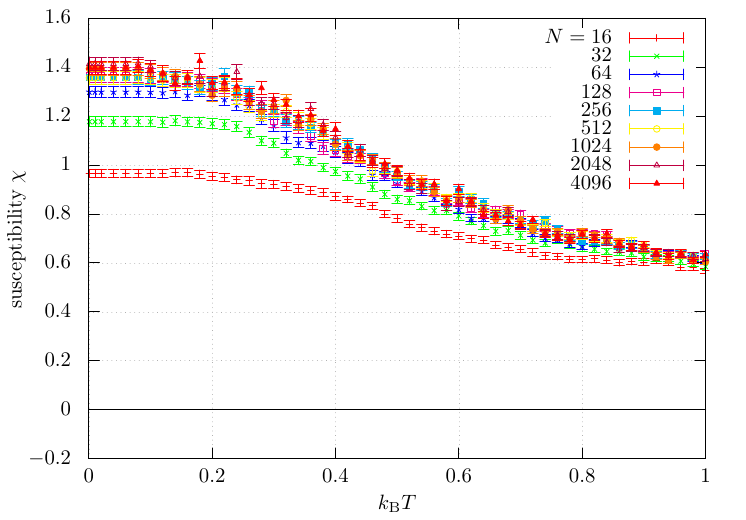}
	\caption{Temperature dependence of the susceptibility, Eq.~\eqref{main_eq_susceptibility_001_001}. The first panel is the result for $t = 0.3$, the second panel is the result for $t = 0.5$, and the third panel is the result for $t = 0.7$. We set $K = 2$, $J = 1.0$, $q = 10^{-1.0}$, and $\epsilon = 0.00$. We show the results for various system sizes $N = 16, 32, \dots, 4096$, and the curves for different $N$ are overlaid in each panel.}
	\label{appendix_fig_sus_K = 2_R=1_q_1.0_e=0.00_t!=0.0_001_001}
\end{figure}

\begin{figure}[t]
	\centering
	\includegraphics[scale=0.54]{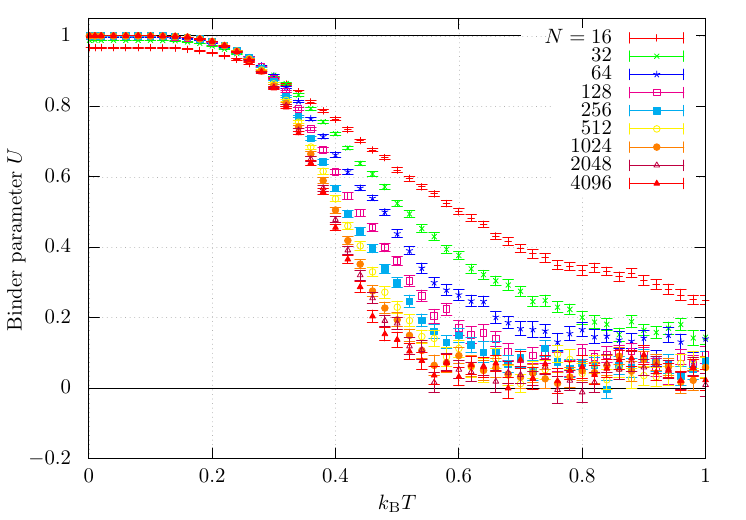}
	\includegraphics[scale=0.54]{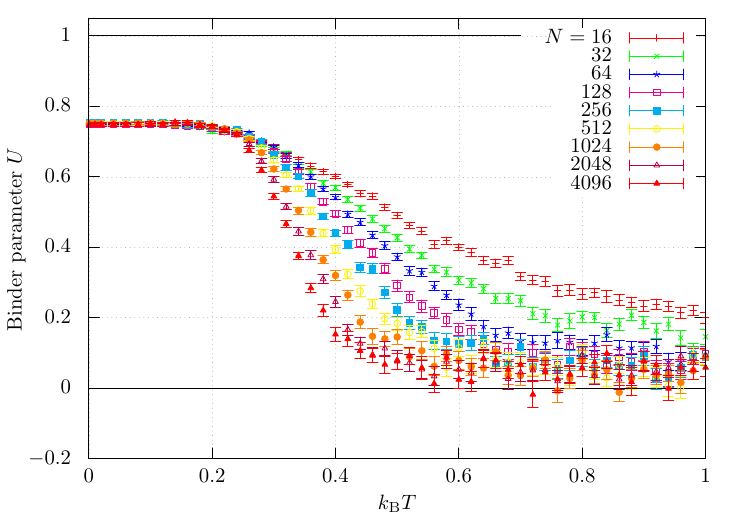}
	\includegraphics[scale=0.54]{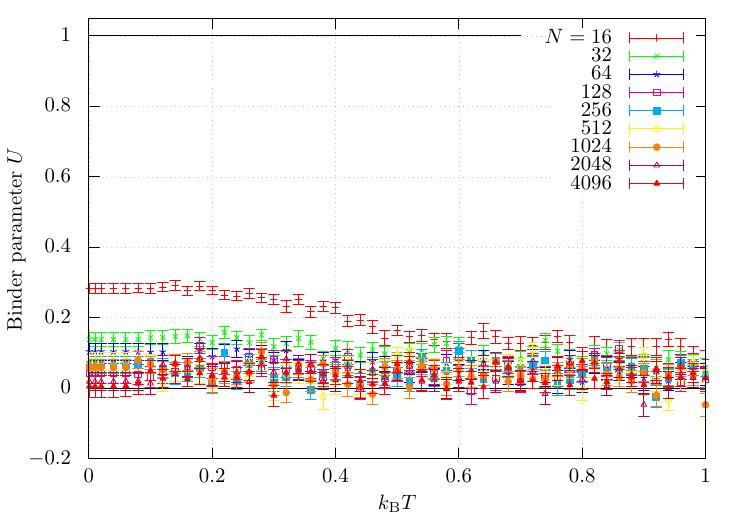}
	\caption{Temperature dependence of the Binder parameter, Eq.~\eqref{main_eq_Binder_parameter_001_001}. The first panel is the result for $t = 0.3$, the second panel is the result for $t = 0.5$, and the third panel is the result for $t = 0.7$. We set $K = 2$, $J = 1.0$, $q = 10^{-2.0}$, and $\epsilon = 0.00$. We show the results for various system sizes $N = 16, 32, \dots, 4096$, and the curves for different $N$ are overlaid in each panel.}
	\label{appendix_fig_Binder_K = 2_R=1_q_2.0_e=0.00_t!=0.0_001_001}
\end{figure}

\begin{figure}[t]
	\centering
	\includegraphics[scale=0.54]{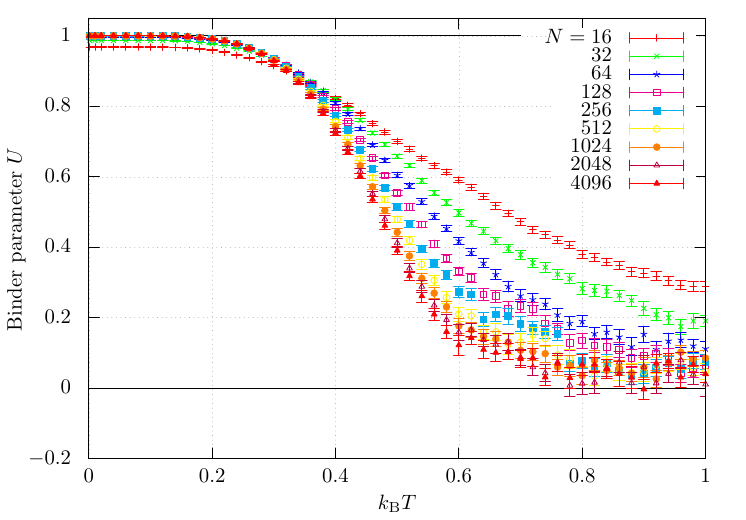}
	\includegraphics[scale=0.54]{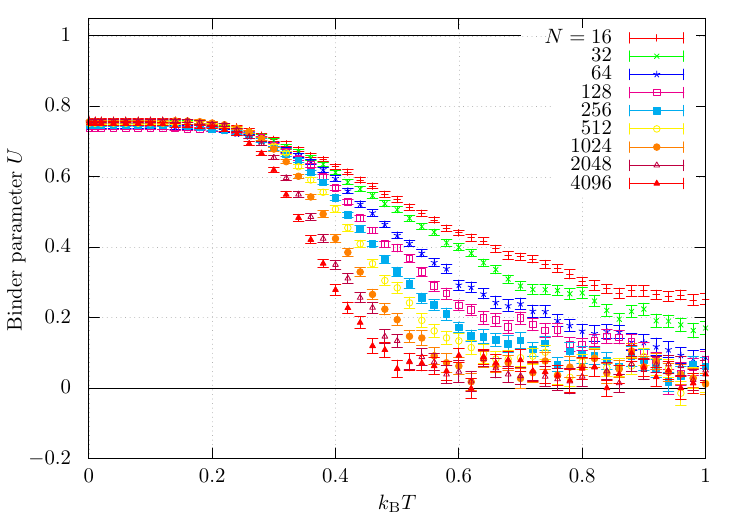}
	\includegraphics[scale=0.54]{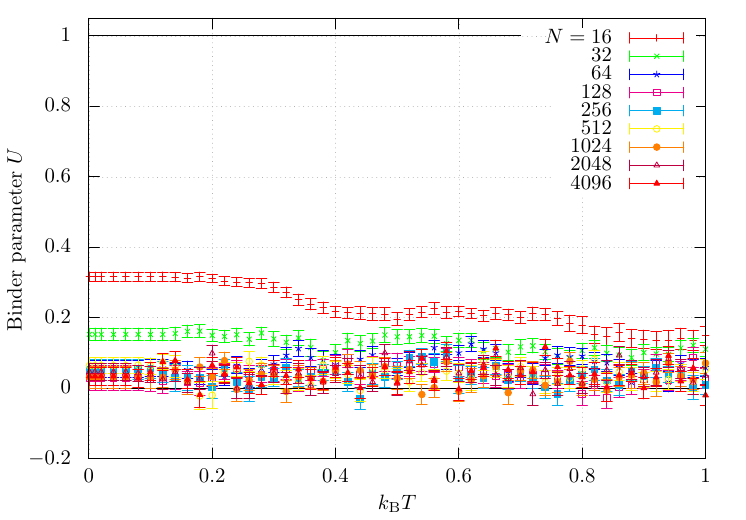}
	\caption{Temperature dependence of the Binder parameter, Eq.~\eqref{main_eq_Binder_parameter_001_001}. The first panel is the result for $t = 0.3$, the second panel is the result for $t = 0.5$, and the third panel is the result for $t = 0.7$. We set $K = 2$, $J = 1.0$, $q = 10^{-1.0}$, and $\epsilon = 0.00$. We show the results for various system sizes $N = 16, 32, \dots, 4096$, and the curves for different $N$ are overlaid in each panel.}
	\label{appendix_fig_Binder_K = 2_R=1_q_1.0_e=0.00_t!=0.0_001_001}
\end{figure}

Below, we compare the results for $t = 0.3, 0.5, 0.7$ and for $q = 10^{-2.0}, 10^{-1.0}$.
The transition temperatures are estimated to be those values shown in the Table~\ref{appendix_table_critical_temperature_K_02_001_001}.
\begin{table}[t]
	\caption{\label{appendix_table_critical_temperature_K_02_001_001}
		The critical temperature of each parameter set.
		The symbol `-' means no phase transitions occur in the parameter setting.}
	\begin{ruledtabular}
		\begin{tabular}{rccc}
			$t$           & 0.3  & 0.5  & 0.7  \\
			\hline
			$q=10^{-2.0}$ & 0.30 & 0.26 & \text{-} \\
			$q=10^{-1.0}$ & 0.36 & 0.28 & \text{-} \\
		\end{tabular}
	\end{ruledtabular}
\end{table}
The magnetization is shown in Figs.~\ref{appendix_fig_mag_K = 2_R=1_q_2.0_e=0.00_t!=0.0_001_001} and \ref{appendix_fig_mag_K = 2_R=1_q_1.0_e=0.00_t!=0.0_001_001}, the susceptibility is shown in Figs.~\ref{appendix_fig_sus_K = 2_R=1_q_2.0_e=0.00_t!=0.0_001_001} and \ref{appendix_fig_sus_K = 2_R=1_q_1.0_e=0.00_t!=0.0_001_001}, and the Binder parameter is shown in Figs.~\ref{appendix_fig_Binder_K = 2_R=1_q_2.0_e=0.00_t!=0.0_001_001} and \ref{appendix_fig_Binder_K = 2_R=1_q_1.0_e=0.00_t!=0.0_001_001}.


\subsubsection{$k_{\mathrm{B}} T$--$t$ phase diagram}

\begin{figure}[t]
	\centering
	\includegraphics[scale=0.60]{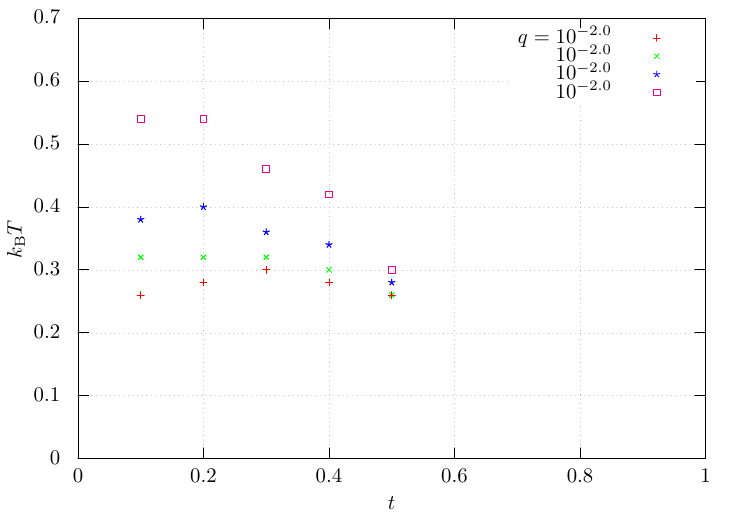}
	\caption{Phase diagram for $K = 2$, $J = 1.0$, and $\epsilon = 0.00$. The horizontal and vertical axes represent the parameter $t$, which controls the probability of selecting production rules, and the critical temperature $k_{\mathrm{B}} T$, respectively.}
	\label{appendix_fig_t_phase_diagram_K = 2_R=1_e=0.00_001_001}
\end{figure}

The $k_{\mathrm{B}} T$--$t$ phase diagram is shown in Fig.~\ref{appendix_fig_t_phase_diagram_K = 2_R=1_e=0.00_001_001}.

\clearpage

\subsection{Case of $K = 20, \epsilon = 0.00$, and $t=0.0$}

\subsubsection{$k_{\mathrm{B}} T$--$q$ phase diagram}

\begin{figure}[t]
	\centering
	\includegraphics[scale=0.54]{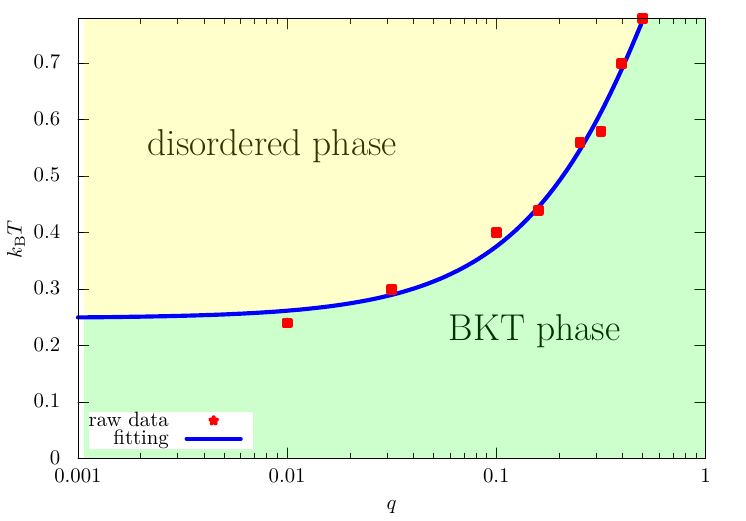}
	\caption{Phase diagram for
		$K = 20$, $J = 1.0$, $t = 0.0$, and $\epsilon = 0.00$.
		The horizontal and vertical axes represent
		the parameter $q$, which controls the probability of selecting production rules,
		and the critical temperature $k_{\mathrm{B}} T$, respectively.}
	\label{appendix_fig_phase_diagram_K = 20_R=1_e=0.00_001_001}
\end{figure}

In Fig.~\ref{appendix_fig_phase_diagram_K = 20_R=1_e=0.00_001_001}, we show the $k_{\mathrm{B}} T$--$q$ phase diagram of $K = 20$ and $\epsilon = 0.00$.

\subsubsection{Magnetization, susceptibility and Binder parameter}

\begin{figure}[t]
	\centering
	\includegraphics[scale=0.54]{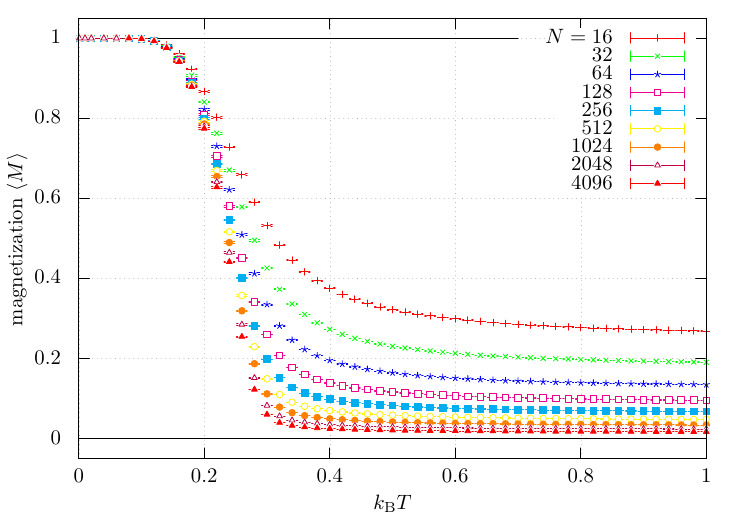}
	\includegraphics[scale=0.54]{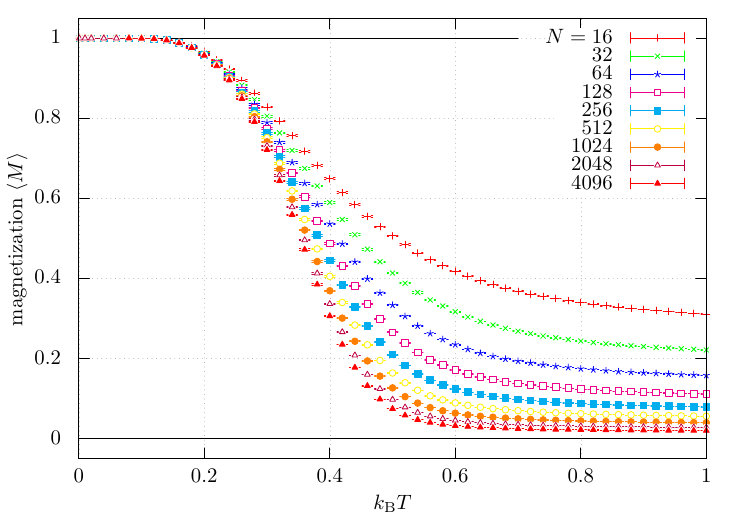}
	\caption{Temperature dependence of the magnetization, Eq.~\eqref{main_eq_magnetization_001_001}. The top panel is the result for $q = 10^{-2.0}$, and the bottom panel is the result for $q = 10^{-1.0}$. We set $K = 20$, $J = 1.0$, $t = 0.0$, and $\epsilon = 0.00$. We show the results for various system sizes $N = 16, 32, \dots, 4096$, and the curves for different $N$ are overlaid in each panel.}
	\label{appendix_fig_mag_K = 20_R=1_e=0.00_001_001}
\end{figure}

\begin{figure}[t]
	\centering
	\includegraphics[scale=0.54]{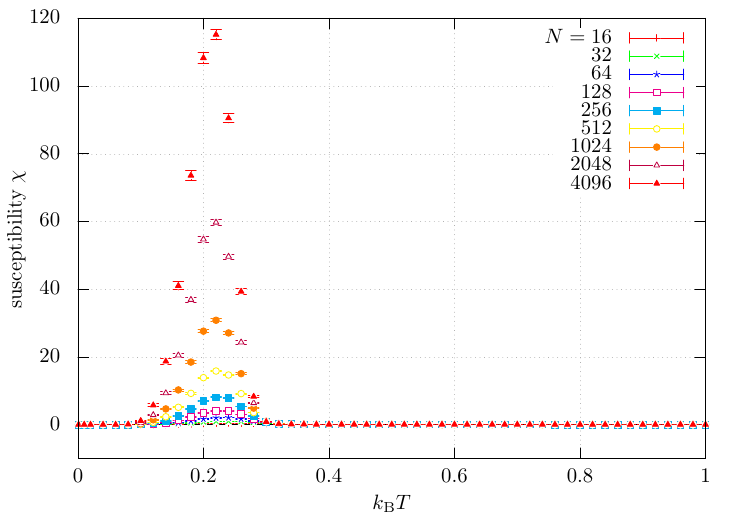}
	\includegraphics[scale=0.54]{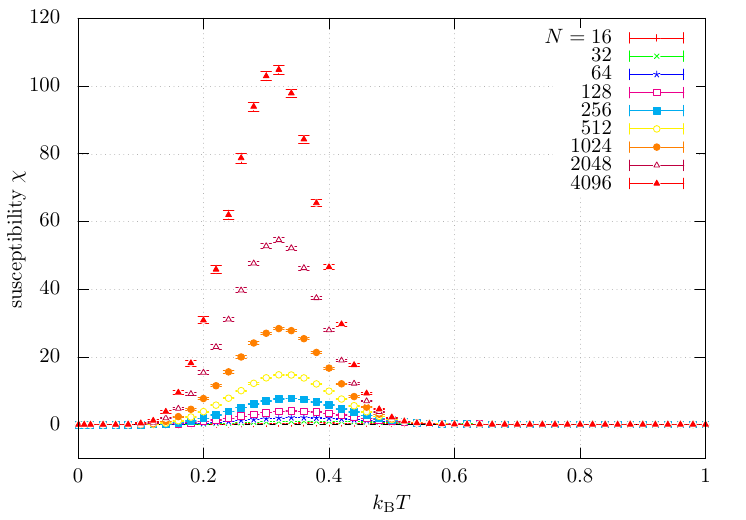}
	\caption{Temperature dependence of the susceptibility, Eq.~\eqref{main_eq_susceptibility_001_001}. The top panel is the result for $q = 10^{-2.0}$, and the bottom panel is the result for $q = 10^{-1.0}$. We set $K = 20$, $J = 1.0$, $t = 0.0$, and $\epsilon = 0.00$. We show the results for various system sizes $N = 16, 32, \dots, 4096$, and the curves for different $N$ are overlaid in each panel.}
	\label{appendix_fig_sus_K = 20_R=1_e=0.00_001_001}
\end{figure}

\begin{figure}[t]
	\centering
	\includegraphics[scale=0.54]{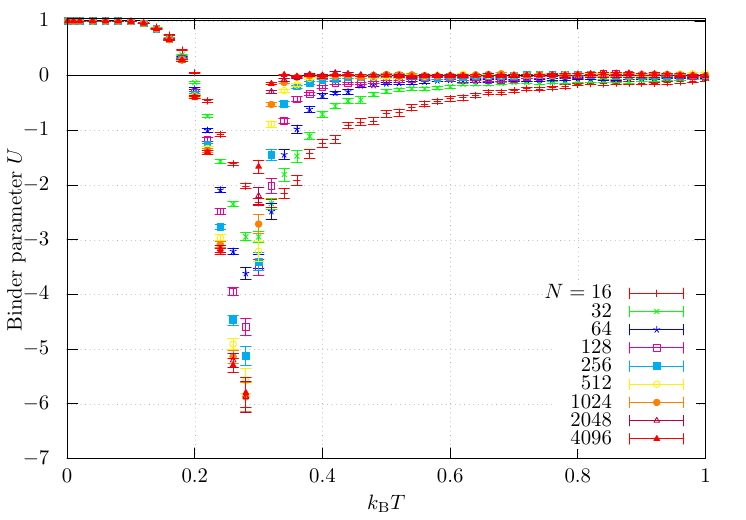}
	\includegraphics[scale=0.54]{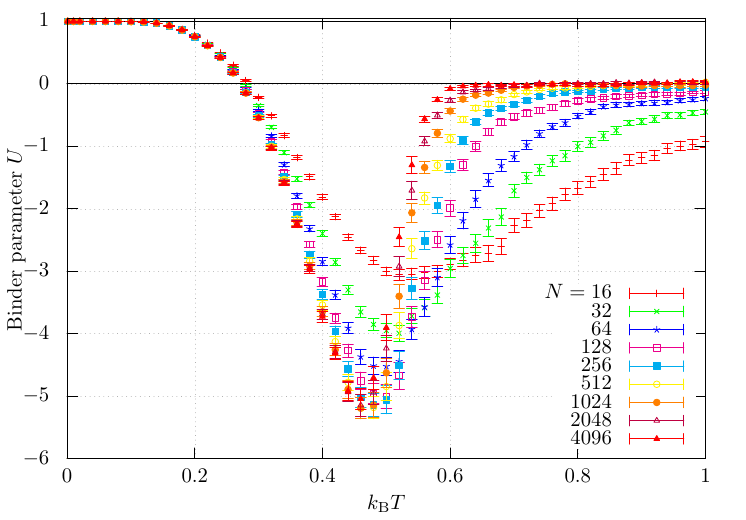}
	\caption{Temperature dependence of the Binder parameter, Eq.~\eqref{main_eq_Binder_parameter_001_001}. The top panel is the result for $q = 10^{-2.0}$, and the bottom panel is the result for $q = 10^{-1.0}$. We set $K = 20$, $J = 1.0$, $t = 0.0$, and $\epsilon = 0.00$. We show the results for various system sizes $N = 16, 32, \dots, 4096$, and the curves for different $N$ are overlaid in each panel.}
	\label{appendix_fig_Binder_K = 20_R=1_e=0.00_001_001}
\end{figure}

Below, we compare the results for $q = 10^{-2.0}, 10^{-1.0}$.
The transition temperatures are estimated to be $k_{\mathrm{B}} T = 0.20$ for $q = 10^{-2.0}$ and $k_{\mathrm{B}} T = 0.34$ for $q = 10^{-1.0}$.
The magnetization is shown in Fig.~\ref{appendix_fig_mag_K = 20_R=1_e=0.00_001_001}, the susceptibility is shown in Fig.~\ref{appendix_fig_sus_K = 20_R=1_e=0.00_001_001}, and the Binder parameter is shown in Fig.~\ref{appendix_fig_Binder_K = 20_R=1_e=0.00_001_001}.


We also show the system-size dependence of the magnetization, susceptibility, and Binder parameter.
Figure~\ref{appendix_fig_mag_K = 20_R=1_e=0.00_001_001} shows the system-size dependence of the magnetization, Fig.~\ref{appendix_fig_sus_K = 20_R=1_e=0.00_001_001} shows that of the susceptibility, and Fig.~\ref{appendix_fig_Binder_K = 20_R=1_e=0.00_001_001} shows that of the Binder parameter.
\begin{figure}[t]
	\centering
	\includegraphics[scale=0.54]{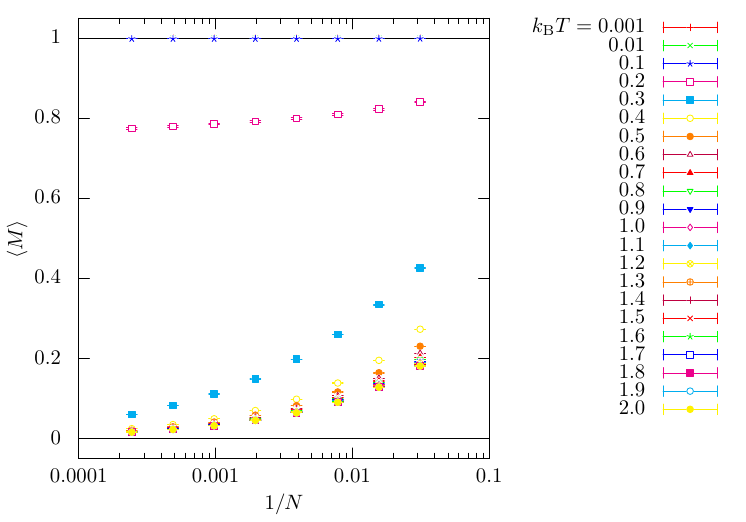}
	\includegraphics[scale=0.54]{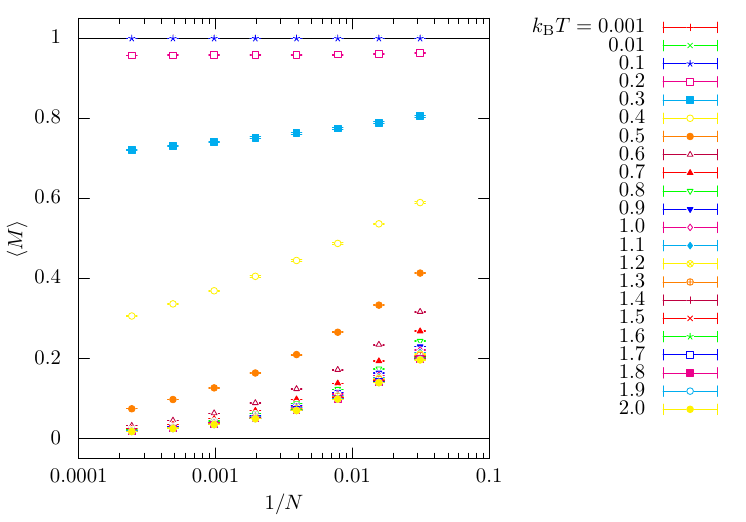}
	\caption{The system-size dependence of the magnetization, Eq.~\eqref{main_eq_magnetization_001_001}. The top panel is the result for $q = 10^{-2.0}$, and the bottom panel is the result for $q = 10^{-1.0}$. We set $K = 20$, $J = 1.0$, $t = 0.0$, and $\epsilon = 0.00$. We show the results for various $k_{\mathrm{B}} T = 0.001, 0.01, 0.1, 0.2, \dots, 2.0$, and the curves for different $k_{\mathrm{B}} T$ are overlaid.}
	\label{appendix_fig_size_dependence_mag_K = 20_R=1_e=0.00_001_001}
\end{figure}
\begin{figure}[t]
	\centering
	\includegraphics[scale=0.54]{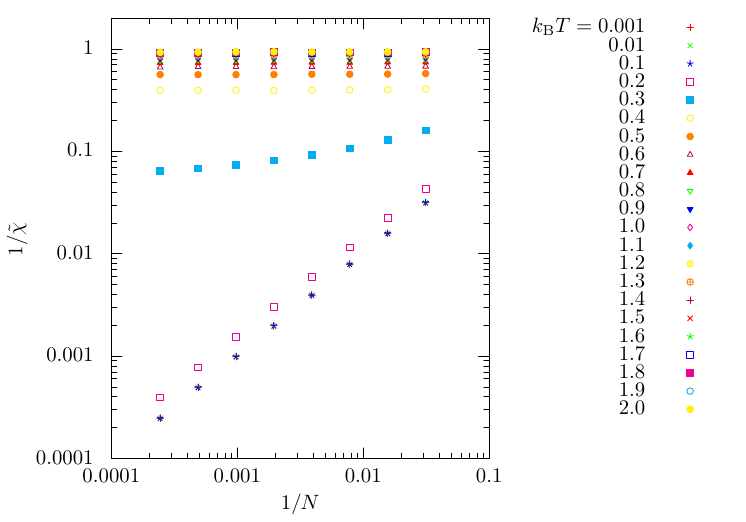}
	\includegraphics[scale=0.54]{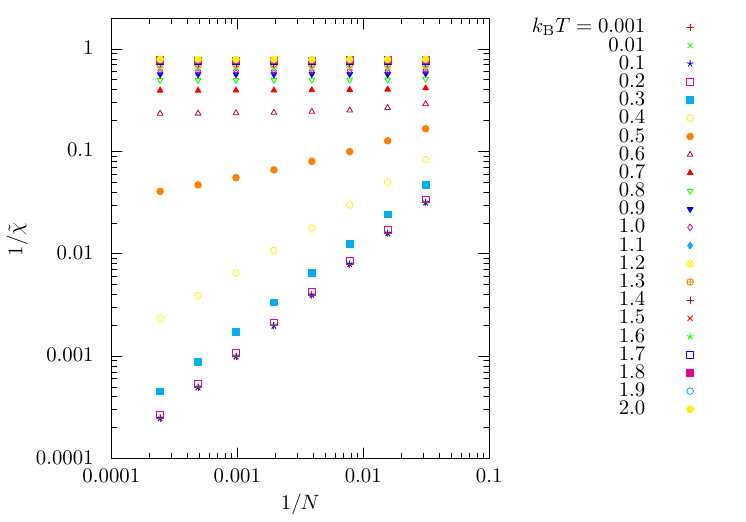}
	\caption{The system-size dependence of the susceptibility, Eq.~\eqref{main_eq_susceptibility_001_001}. The top panel is the result for $q = 10^{-2.0}$, and the bottom panel is the result for $q = 10^{-1.0}$. We set $K = 20$, $J = 1.0$, $t = 0.0$, and $\epsilon = 0.00$. We show the results for various $k_{\mathrm{B}} T = 0.001, 0.01, 0.1, 0.2, \dots, 2.0$, and the curves for different $k_{\mathrm{B}} T$ are overlaid.}
	\label{appendix_fig_size_dependence_sus_K = 20_R=1_e=0.00_001_001}
\end{figure}
\begin{figure}[t]
	\centering
	\includegraphics[scale=0.54]{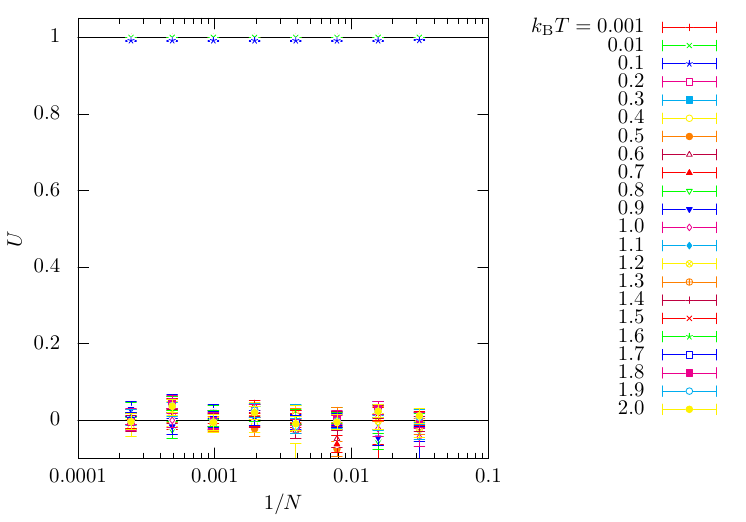}
	\includegraphics[scale=0.54]{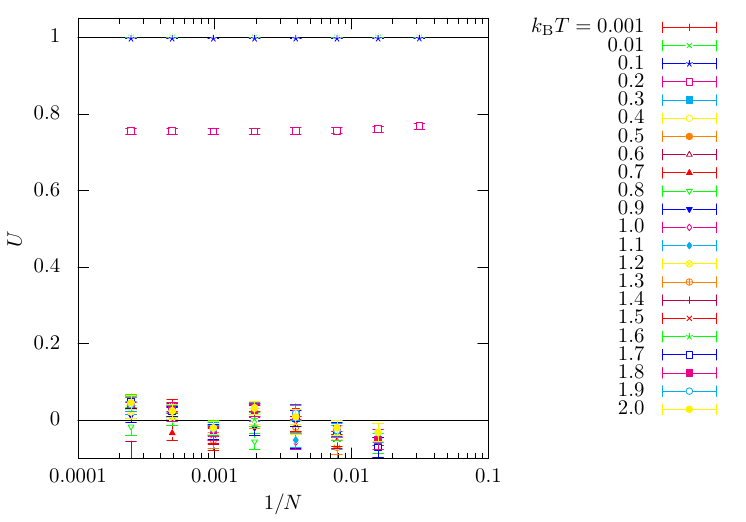}
	\caption{The system-size dependence of the Binder parameter, Eq.~\eqref{main_eq_Binder_parameter_001_001}. The top panel is the result for $q = 10^{-2.0}$, and the bottom panel is the result for $q = 10^{-1.0}$. We set $K = 20$, $J = 1.0$, $t = 0.0$, and $\epsilon = 0.00$. We show the results for various $k_{\mathrm{B}} T = 0.001, 0.01, 0.1, 0.2, \dots, 2.0$, and the curves for different $k_{\mathrm{B}} T$ are overlaid.}
	\label{appendix_fig_size_dependence_Binder_K = 20_R=1_e=0.00_001_001}
\end{figure}

\subsubsection{Histogram of the magnetization} \label{appendix_subsec_histogram_K = 20_R=1_e=0.00_001}

\begin{figure}[t]
	\centering
	\includegraphics[scale=0.54]{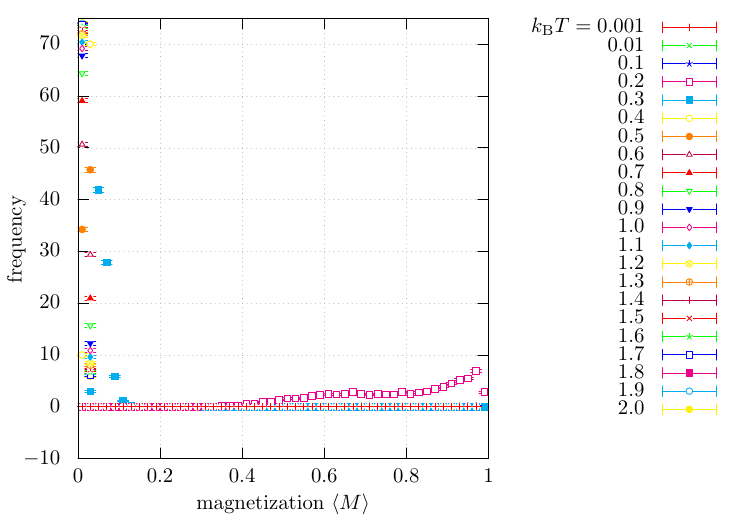}
	\includegraphics[scale=0.54]{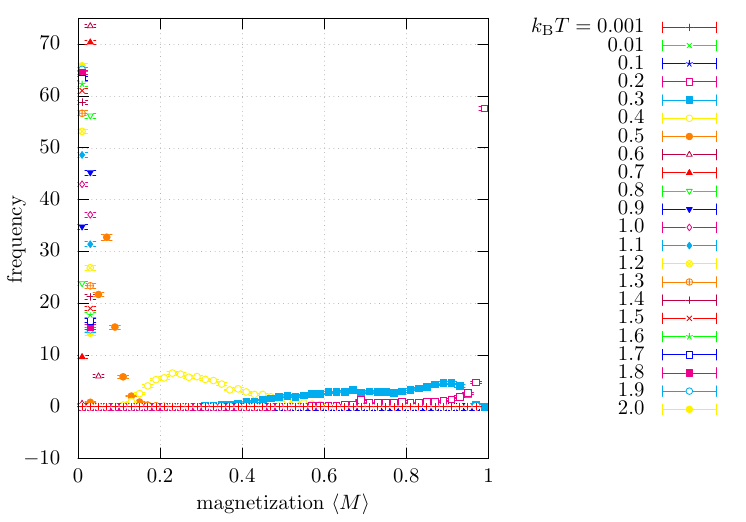}
	\caption{The histogram of the magnetization. The top panel is the result for $q = 10^{-2.0}$, and the bottom panel is the result for $q = 10^{-1.0}$. We set $K = 20$, $J = 1.0$, $t = 0.0$, and $\epsilon = 0.00$. We show the results for various $k_{\mathrm{B}} T = 0.001, 0.1, \dots, 2.0$, and the curves for different $k_{\mathrm{B}} T$ are overlaid.}
	\label{appendix_fig_histogram_K = 20_R=1_e=0.00_001_001}
\end{figure}

The histogram of the magnetization is shown in Fig.~\ref{appendix_fig_histogram_K = 20_R=1_e=0.00_001_001}.

\subsubsection{Typical configurations of symbols}

\begin{figure}[t]
	\centering
	\includegraphics[scale=0.54]{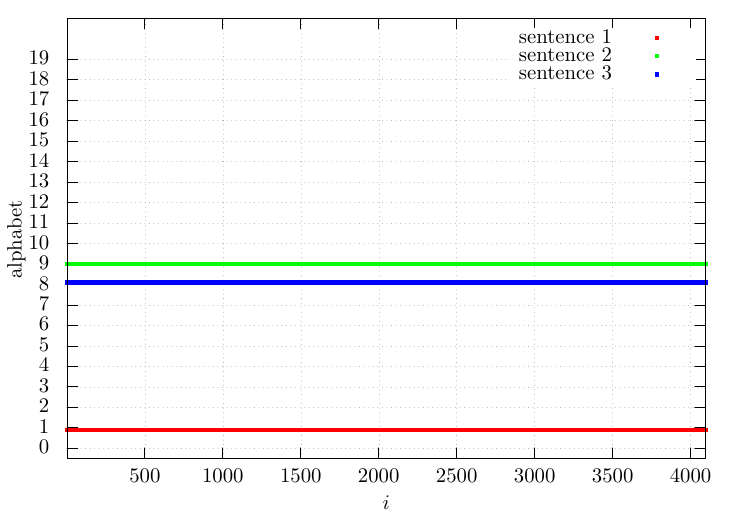}
	\includegraphics[scale=0.54]{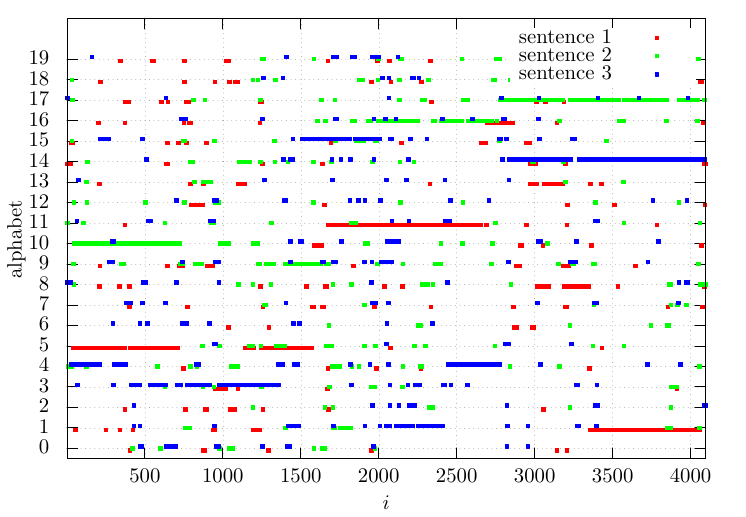}
	\includegraphics[scale=0.54]{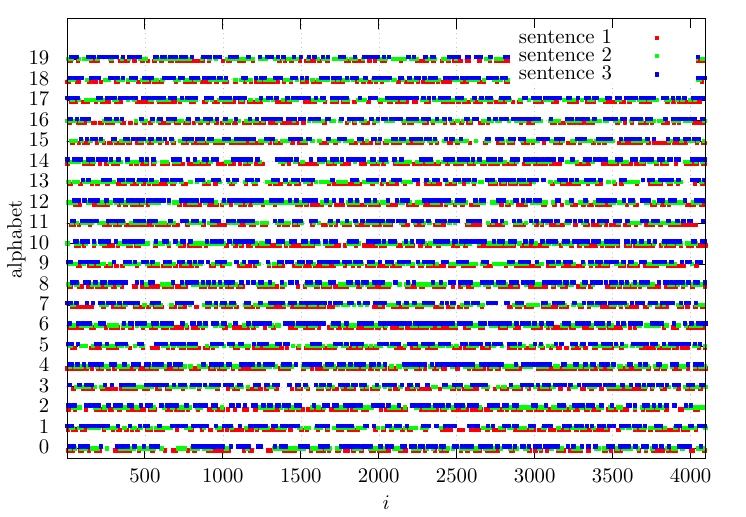}
	\caption{Typical configurations of symbols. The upper panel shows an example sentence at a temperature below the transition ($k_{\mathrm{B}} T = 0.001$) with magnetization $0.98 \leq M < 1.00$. The middle panel shows an example sentence at a temperature just above the transition ($k_{\mathrm{B}} T = 0.20$) with magnetization $0.98 \leq M < 1.00$. The lower panel shows an example sentence at a temperature above the transition ($k_{\mathrm{B}} T = 0.40$) with magnetization $0.00 \leq M < 0.02$. We set $K = 20$, $J = 1.0$, $q = 10^{-2.0}$, $t = 0.0$, and $\epsilon = 0.00$.}
	\label{appendix_fig_configuration_K = 20_q=10_-2.0_R=1_e=0.00_001_001}
\end{figure}

\begin{figure}[t]
	\centering
	\includegraphics[scale=0.54]{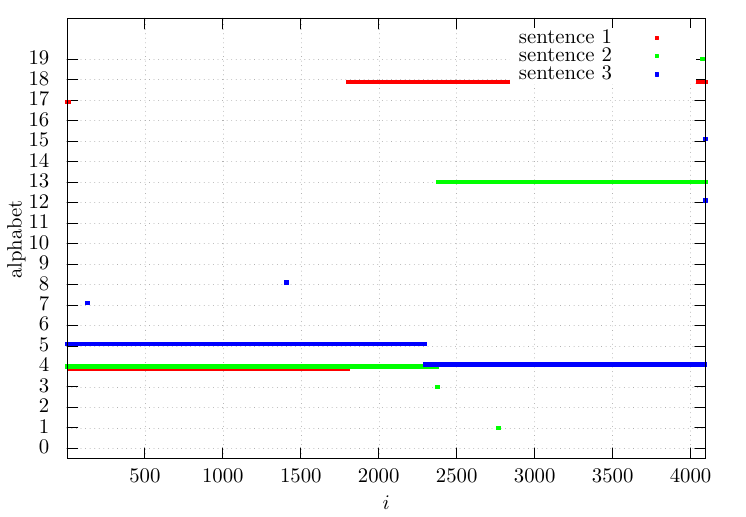}
	\includegraphics[scale=0.54]{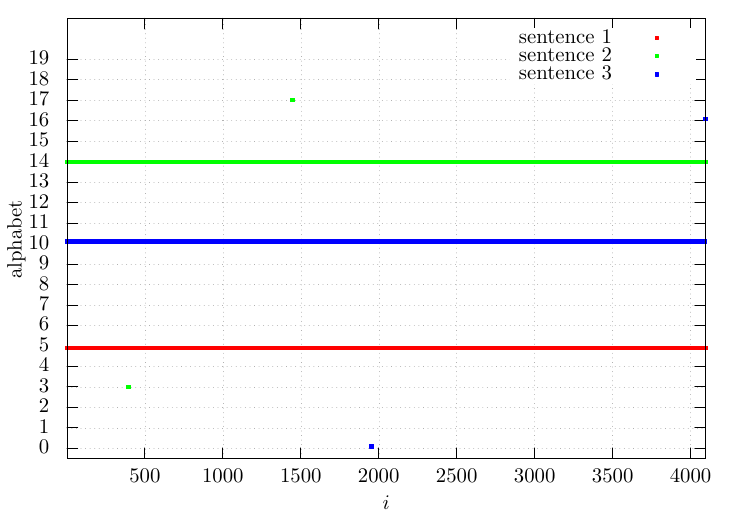}
	\includegraphics[scale=0.54]{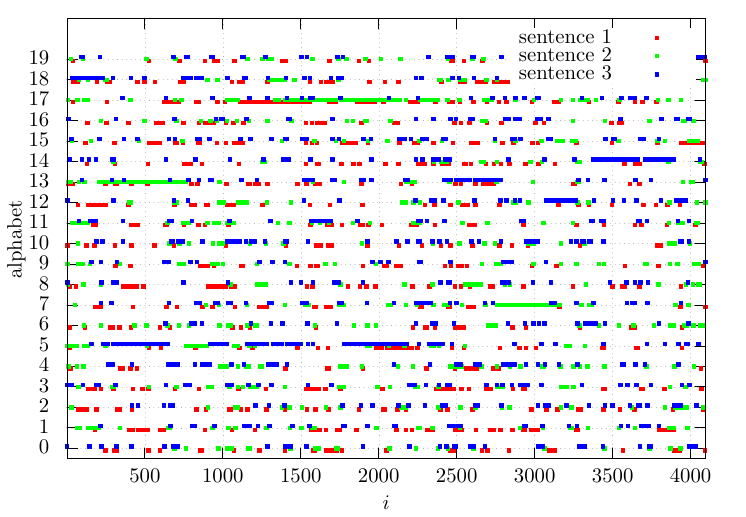}
	\includegraphics[scale=0.54]{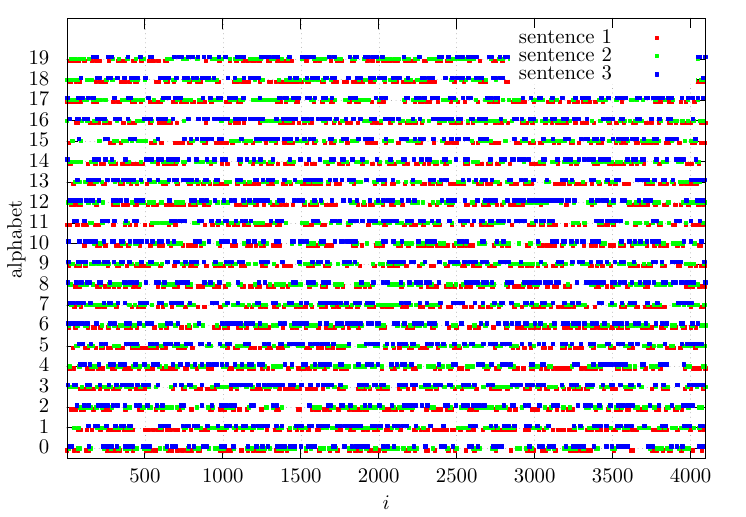}
	\caption{Typical configurations of symbols. The first panel shows an example sentence at a temperature below the transition ($k_{\mathrm{B}} T = 0.20$) with magnetization $0.68 \leq M < 0.70$. The second panel shows an example sentence at the same temperature ($k_{\mathrm{B}} T = 0.20$) with magnetization $0.98 \leq M < 1.00$. The third panel shows an example sentence at a temperature just above the transition ($k_{\mathrm{B}} T = 0.40$) with magnetization $0.22 \leq M < 0.24$. The fourth panel shows an example sentence at a temperature above the transition ($k_{\mathrm{B}} T = 0.60$) with magnetization $0.02 \leq M < 0.04$. We set $K = 20, R_{-} = R_{+} = 1, J = 0, q = 10^{-1.0}, t = 0.0, s = 0.9$, and $\epsilon = 0.00$. The first and second panels correspond to the same value of $k_{\mathrm{B}} T$, because the histogram at this temperature exhibits two peaks, and we show configurations corresponding to both peaks. However, the configurations with $0.98 \leq M < 1.00$ appear more frequently than those with $0.68 \leq M < 0.70$.}
	\label{appendix_fig_configuration_K = 20_q=10_-1.0_R=1_e=0.00_001_001}
\end{figure}

By examining the histograms shown in Sec.~\ref{appendix_subsec_histogram_K = 20_R=1_e=0.00_001}, we can identify, for each temperature, the most frequent value of the magnetization.
We regard this value as the typical magnetization at that temperature, and we show, as examples, the configurations of the generated text whose magnetization takes this typical value in Figs.~\ref{appendix_fig_configuration_K = 20_q=10_-2.0_R=1_e=0.00_001_001} and \ref{appendix_fig_configuration_K = 20_q=10_-1.0_R=1_e=0.00_001_001}.
The temperatures shown correspond to values below the transition temperature, just above the transition temperature, and above the transition temperature.

\subsubsection{Correlation function}

\begin{figure}[t]
	\centering
	\includegraphics[scale=0.54]{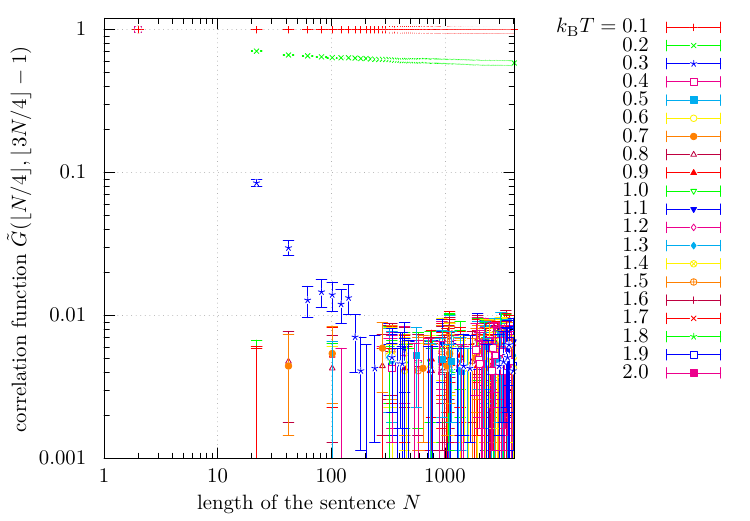}
	\includegraphics[scale=0.54]{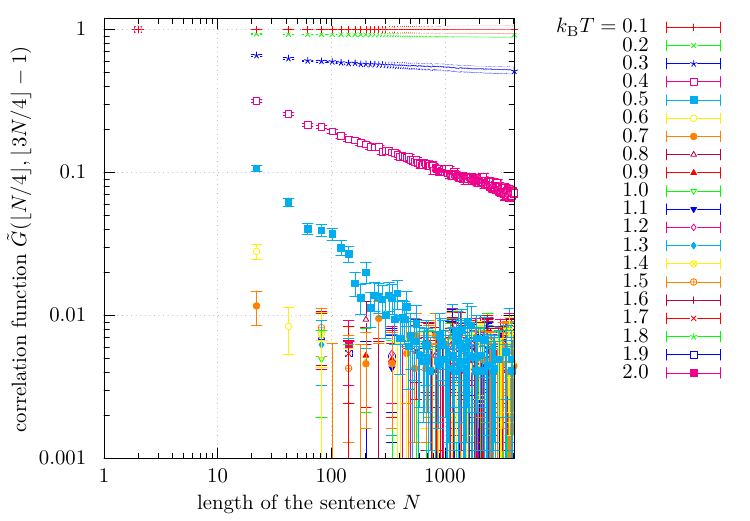}
	\caption{The correlation function, Eq.~\eqref{main_eq_correlation_function_001_002}, with $i = \lfloor N/4 \rfloor$ and $j = \lfloor 3N/4 \rfloor - 1$. The top panel is the result for $q = 10^{-2.0}$, and the bottom panel is the result for $q = 10^{-1.0}$. We set $K = 20$, $J = 1.0$, $t = 0.0$, and $\epsilon = 0.00$. We show the results for various $k_{\mathrm{B}} T = 0.1, 0.2, \dots, 2.0$, and the curves for different $k_{\mathrm{B}} T$ are overlaid.}
	\label{appendix_fig_correlation_K = 20_R=1_e=0.00_001_001}
\end{figure}

The data of the correlation function $\tilde{G}(\lfloor N/4 \rfloor, \lfloor 3N/4 \rfloor - 1)$ are shown in Fig.~\ref{appendix_fig_correlation_K = 20_R=1_e=0.00_001_001}.

\subsubsection{Mutual information}

\begin{figure}[t]
	\centering
	\includegraphics[scale=0.54]{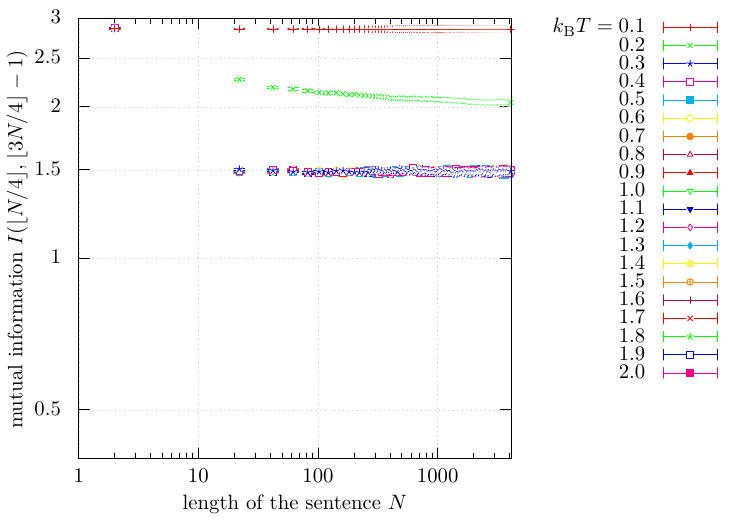}
	\includegraphics[scale=0.54]{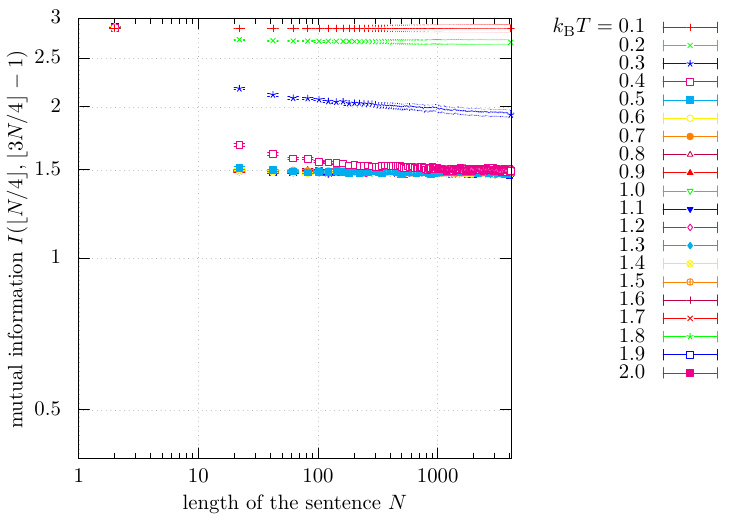}
	\caption{The mutual information $I(\lfloor N/4 \rfloor, \lfloor 3N/4 \rfloor - 1)$. The top panel is the result for $q = 10^{-2.0}$, and the bottom panel is the result for $q = 10^{-1.0}$. We set $K = 20$, $J = 1.0$, $t = 0.0$, and $\epsilon = 0.00$. We show the results for various $k_{\mathrm{B}} T = 0.1, 0.2, \dots, 2.0$, and the curves for different $k_{\mathrm{B}} T$ are overlaid.}
	\label{appendix_fig_mutual_information_K = 20_R=1_e=0.00_001_001}
\end{figure}

The data of the mutual information $I(\lfloor N/4 \rfloor, \lfloor 3N/4 \rfloor - 1)$ are shown in Fig.~\ref{appendix_fig_mutual_information_K = 20_R=1_e=0.00_001_001}.

\subsubsection{Finite-size scaling}

\begin{figure}[t]
	\centering
	\includegraphics[scale=0.54]{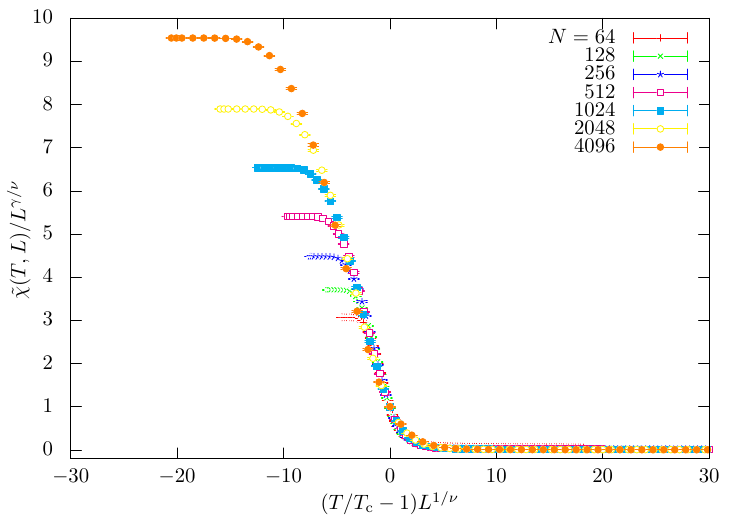}
	\includegraphics[scale=0.54]{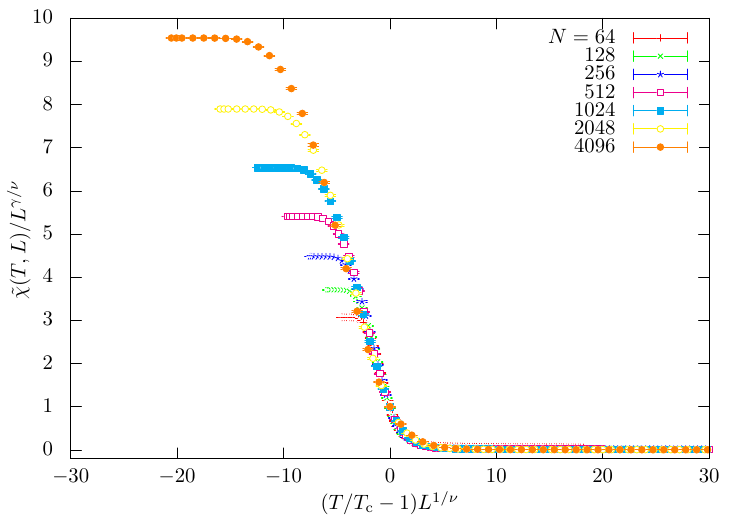}
	\caption{The finite-size scaling plot. The top panel is the result for $q = 10^{-2.0}$, and the bottom panel is the result for $q = 10^{-1.0}$. We plot $\tilde{\chi}(T,N) / N^{\gamma/\nu}$ versus $N^{1/\nu} t$. The critical exponents are set to $\nu = 2.50$ and $\gamma = 2.00$ for $q = 10^{-2.0}$, and to $\nu = 2.75$ and $\gamma = 2.00$ for $q = 10^{-1.0}$. We set $K = 20$, $J = 1.0$, $t = 0.0$, and $\epsilon = 0.00$. We show the results for $N = 64, 128, 256, 512, 1024, 2048, 4096$, and the curves for different $N$ are overlaid.}
	\label{appendix_fig_finite_size_scaling_K = 20_R=1_e=0.00_001_001}
\end{figure}

\begin{figure}[t]
	\centering
	\includegraphics[scale=0.54]{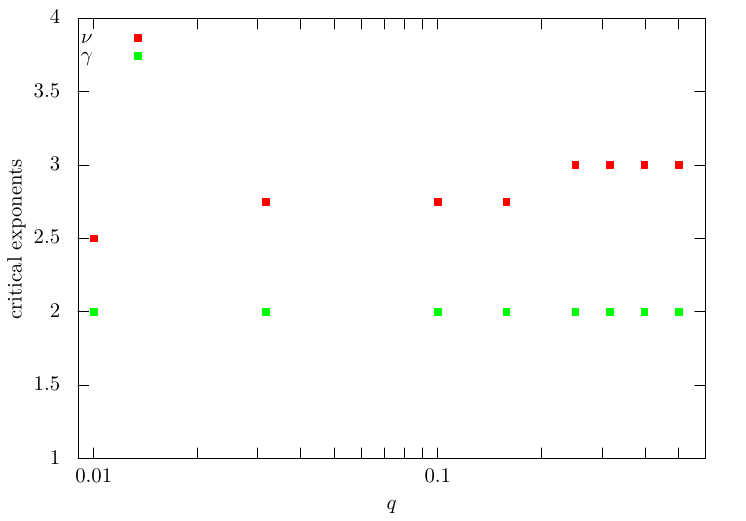}
	\caption{The $q$-dependence of the critical exponents $\nu$ and $\gamma$. We set $K = 20$, $J = 1.0$, $t = 0.0$, and $\epsilon = 0.00$. We show the results for $N = 64, 128, 256, 512, 1024, 2048, 4096$, and the curves for different $N$ are overlaid.}
	\label{appendix_fig_critical_exponents_q_K = 20_R=1_e=0.00_001_001}
\end{figure}

The finite-size scaling plot is shown in Fig.~\ref{appendix_fig_finite_size_scaling_K = 20_R=1_e=0.00_001_001}.
The $q$-dependence of the critical exponents $\gamma$ and $\nu$ is shown in
Fig.~\ref{appendix_fig_critical_exponents_q_K = 20_R=1_e=0.00_001_001}.

\clearpage

\subsection{Case of $K = 200$, $\epsilon = 0.00$, and $t \ne 0$}

\subsubsection{Magnetization, susceptibility and Binder parameter}

\begin{figure}[t]
	\centering
	\includegraphics[scale=0.54]{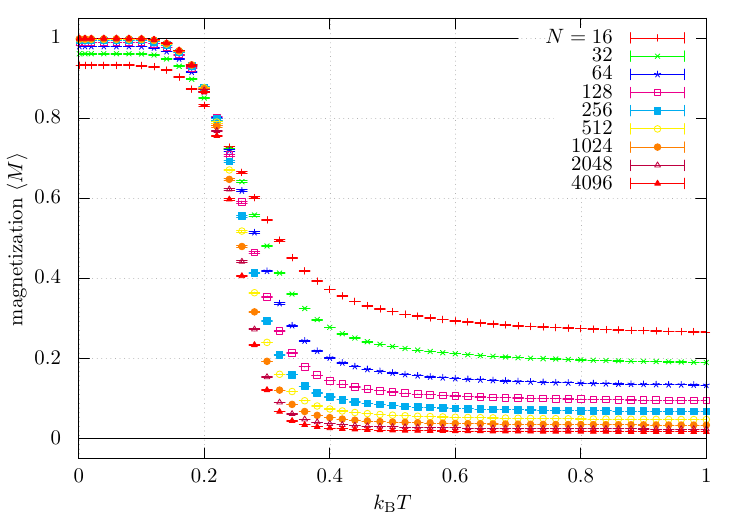}
	\includegraphics[scale=0.54]{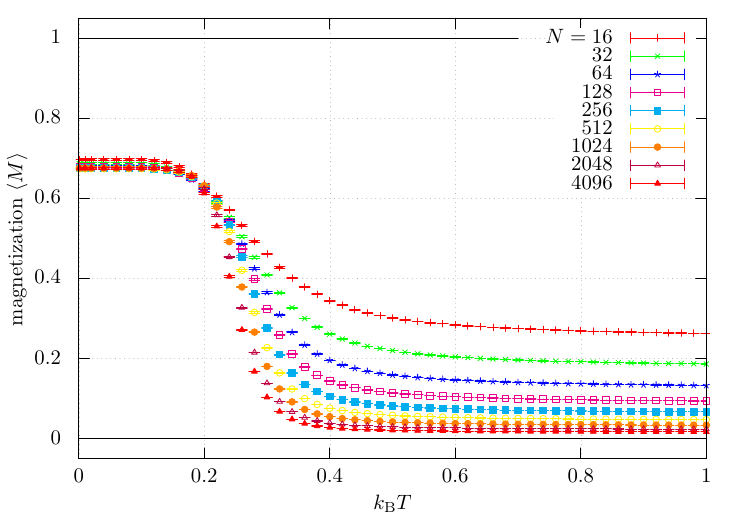}
	\includegraphics[scale=0.54]{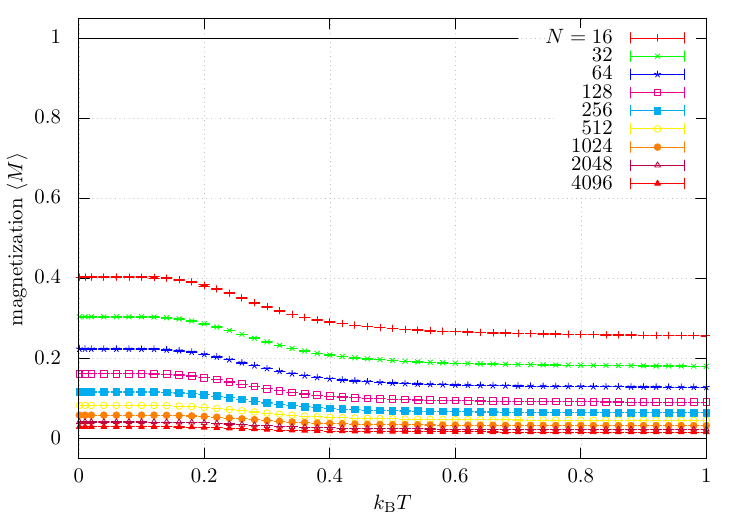}
	\caption{Temperature dependence of the magnetization, Eq.~\eqref{main_eq_magnetization_001_001}. The first panel is the result for $t=0.3$, the second panel is the result for $t=0.5$, and the third panel is the result for $t=0.7$. We set $K = 20, J=1.0,q=10^{-2.0}$, and $\epsilon=0.00$. We show the results for various system sizes $N = 16, 32, \dots, 4096$, and the curves for different $N$ are overlaid in each panel.}
	\label{appendix_fig_mag_K = 20_R=1_q_2.0_e=0.00_t!=0.0_001_001}
\end{figure}

\begin{figure}[t]
	\centering
	\includegraphics[scale=0.54]{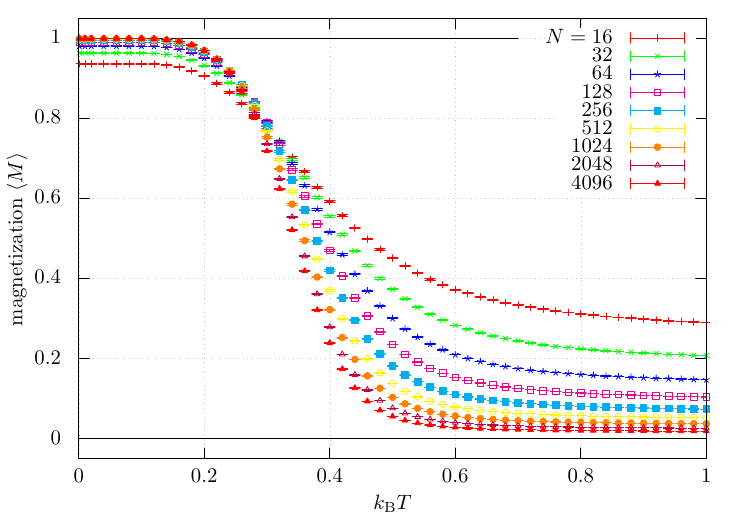}
	\includegraphics[scale=0.54]{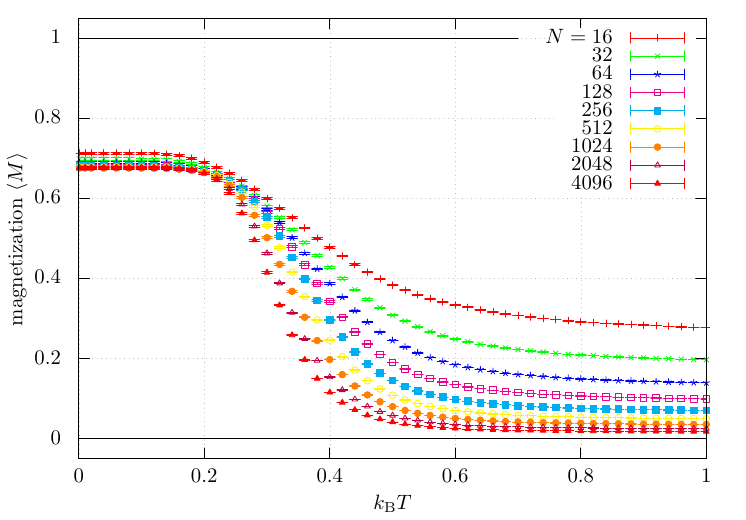}
	\includegraphics[scale=0.54]{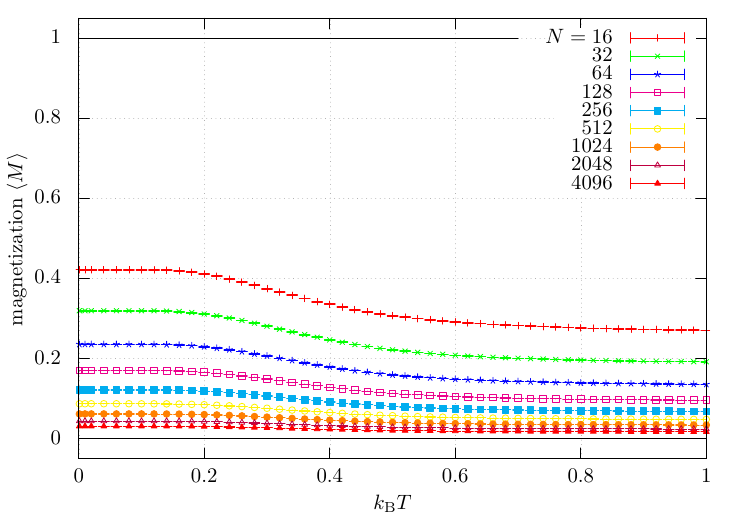}
	\caption{Temperature dependence of the magnetization, Eq.~\eqref{main_eq_magnetization_001_001}.
		The first panel is the result for $t=0.3$, the second panel is the result for $t=0.5$,
		and the third panel is the result for $t=0.7$.
		We set $K = 20$, $J = 1.0$, $q = 10^{-1.0}$, and $\epsilon=0.00$.
		We show the results for various system sizes $N = 16, 32, \dots, 4096$,
		and the curves for different $N$ are overlaid in each panel.}
	\label{appendix_fig_mag_K = 20_R=1_q_1.0_e=0.00_t!=0.0_001_001}
\end{figure}

\begin{figure}[t]
	\centering
	\includegraphics[scale=0.54]{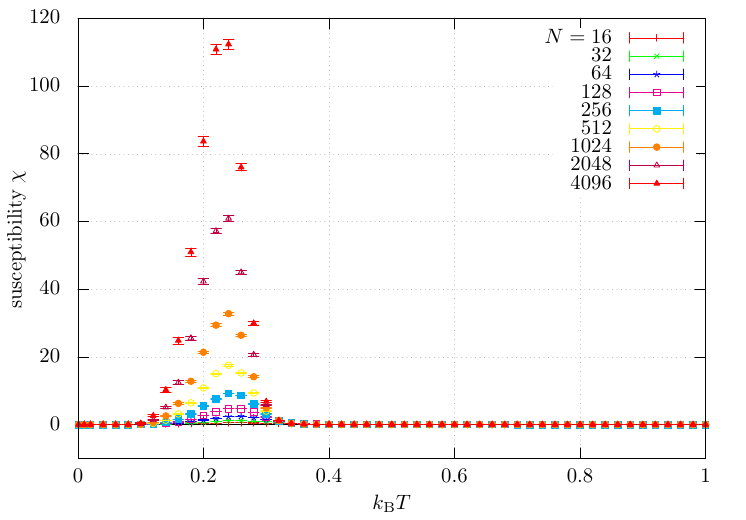}
	\includegraphics[scale=0.54]{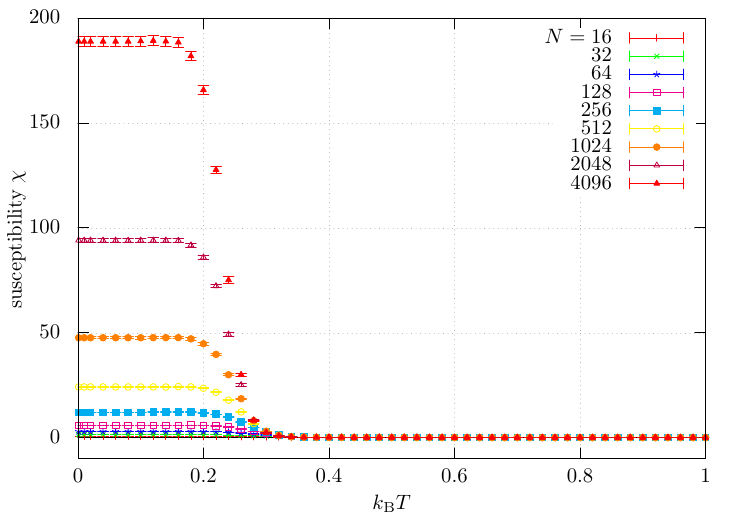}
	\includegraphics[scale=0.54]{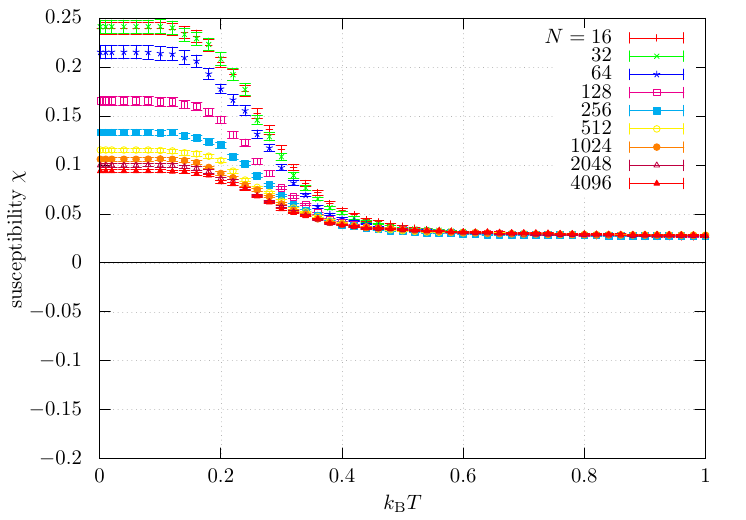}
	\caption{Temperature dependence of the susceptibility, Eq.~\eqref{main_eq_susceptibility_001_001}. The first panel is the result for $t=0.3$, the second panel is the result for $t=0.5$, and the third panel is the result for $t=0.7$. We set $K = 20, J=1.0,q=10^{-2.0}$, and $\epsilon=0.00$. We show the results for various system sizes $N = 16, 32, \dots, 4096$, and the curves for different $N$ are overlaid in each panel.}
	\label{appendix_fig_sus_K = 20_R=1_q_2.0_e=0.00_t!=0.0_001_001}
\end{figure}

\begin{figure}[t]
	\centering
	\includegraphics[scale=0.54]{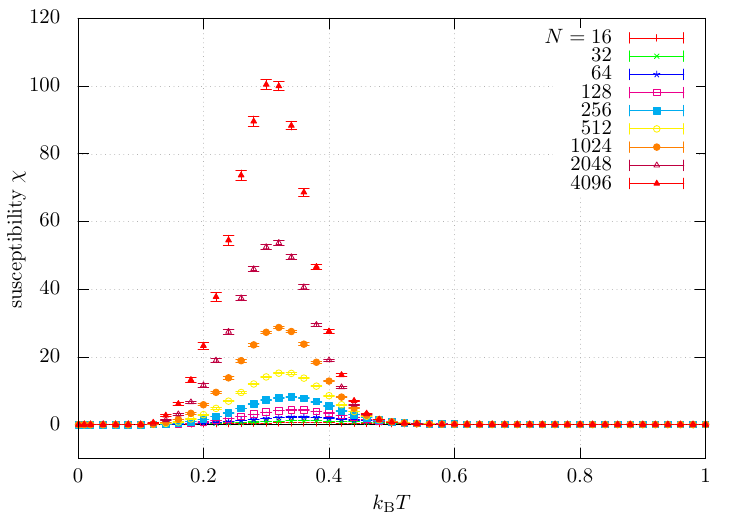}
	\includegraphics[scale=0.54]{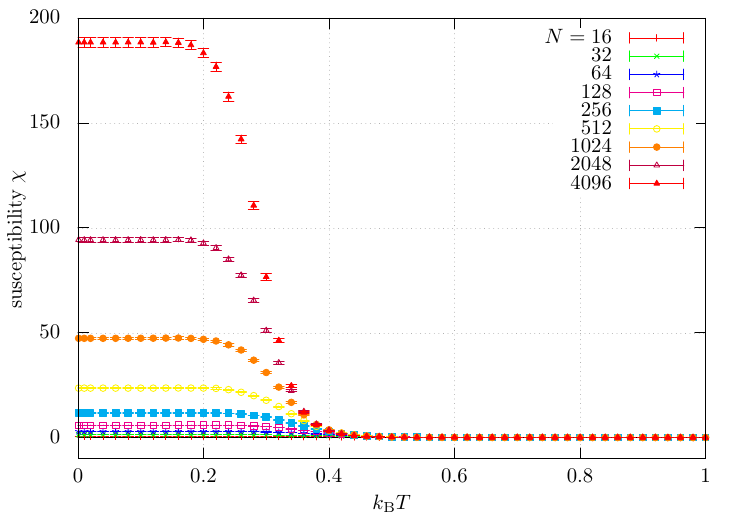}
	\includegraphics[scale=0.54]{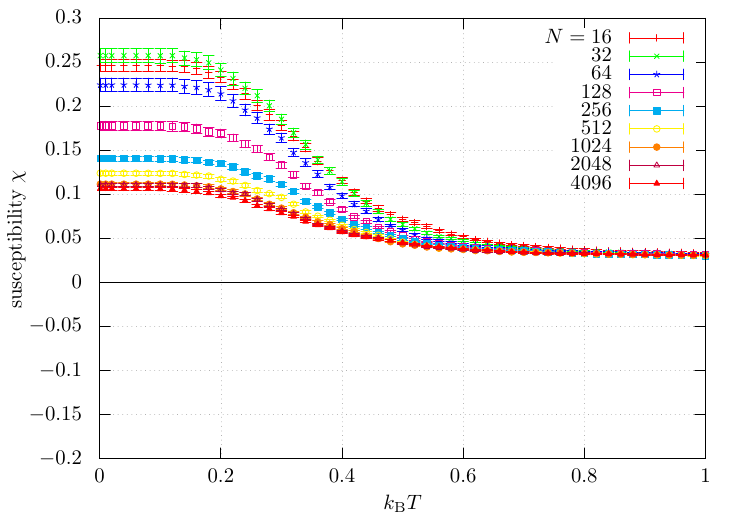}
	\caption{Temperature dependence of the susceptibility, Eq.~\eqref{main_eq_susceptibility_001_001}. The first panel is the result for $t=0.3$, the second panel is the result for $t=0.5$, and the third panel is the result for $t=0.7$. We set $K = 20$, $J = 1.0$, $q = 10^{-1.0}$, and $\epsilon=0.00$. We show the results for various system sizes $N = 16, 32, \dots, 4096$, and the curves for different $N$ are overlaid in each panel.}
	\label{appendix_fig_sus_K = 20_R=1_q_1.0_e=0.00_t!=0.0_001_001}
\end{figure}

\begin{figure}[t]
	\centering
	\includegraphics[scale=0.54]{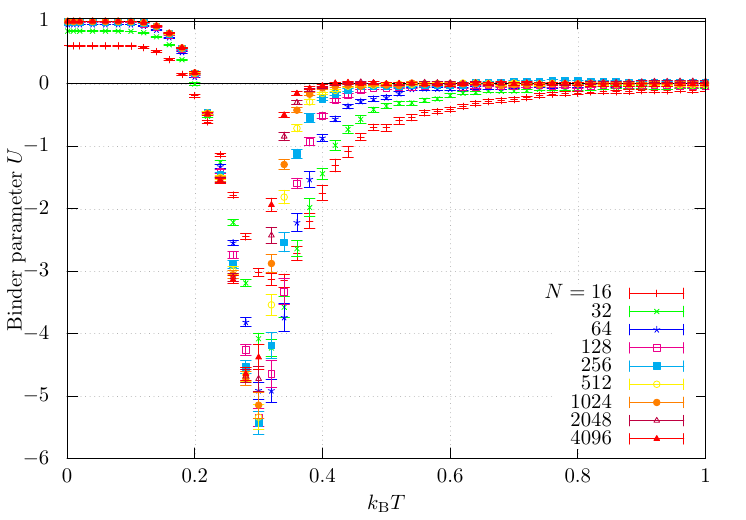}
	\includegraphics[scale=0.54]{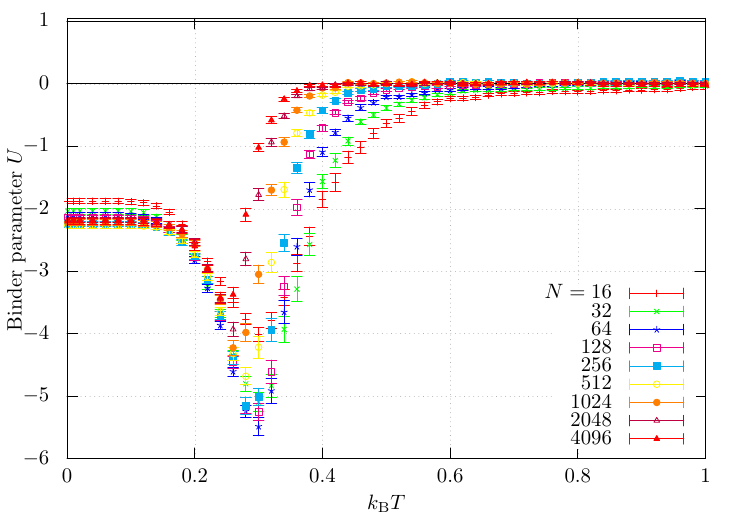}
	\includegraphics[scale=0.54]{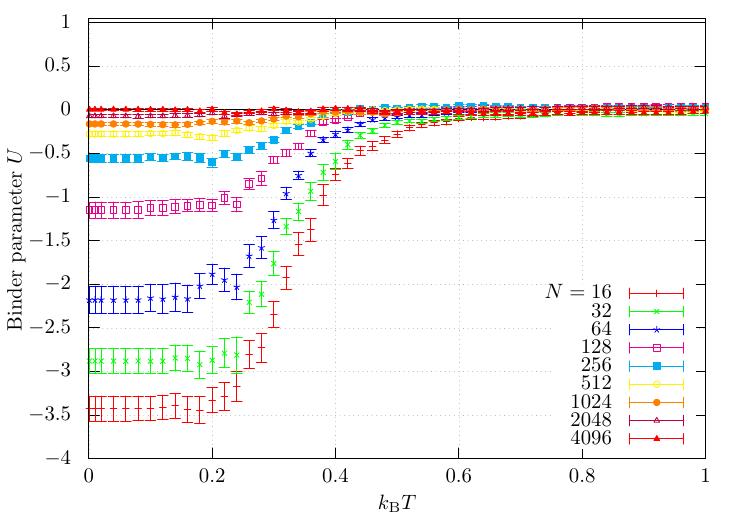}
	\caption{Temperature dependence of the Binder parameter, Eq.~\eqref{main_eq_Binder_parameter_001_001}. The first panel is the result for $t=0.3$, the second panel is the result for $t=0.5$, and the third panel is the result for $t=0.7$. We set $K = 20, J=1.0,q=10^{-2.0}$, and $\epsilon=0.00$. We show the results for various system sizes $N = 16, 32, \dots, 4096$, and the curves for different $N$ are overlaid in each panel.}
	\label{appendix_fig_Binder_K = 20_R=1_q_2.0_e=0.00_t!=0.0_001_001}
\end{figure}

\begin{figure}[t]
	\centering
	\includegraphics[scale=0.54]{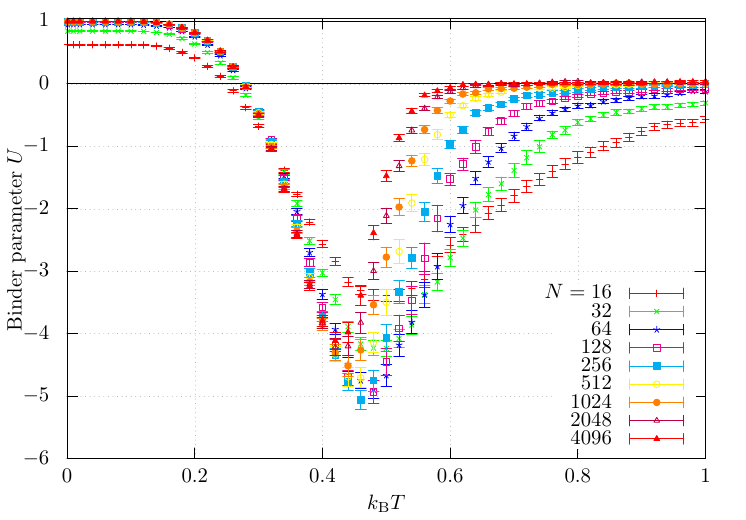}
	\includegraphics[scale=0.54]{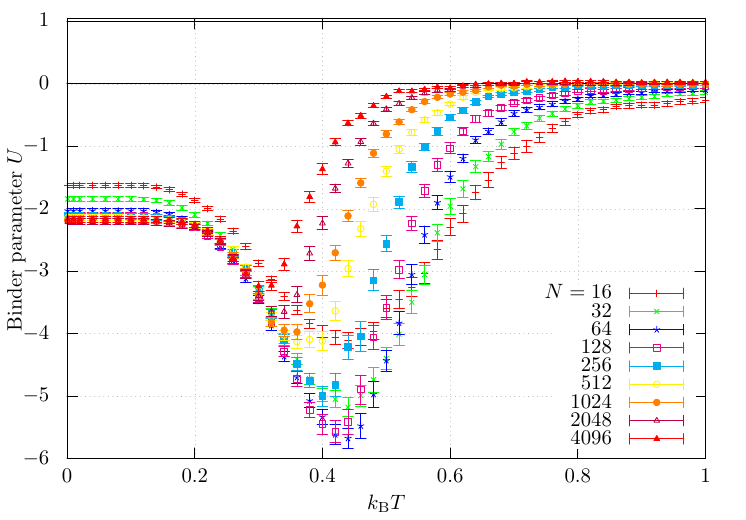}
	\includegraphics[scale=0.54]{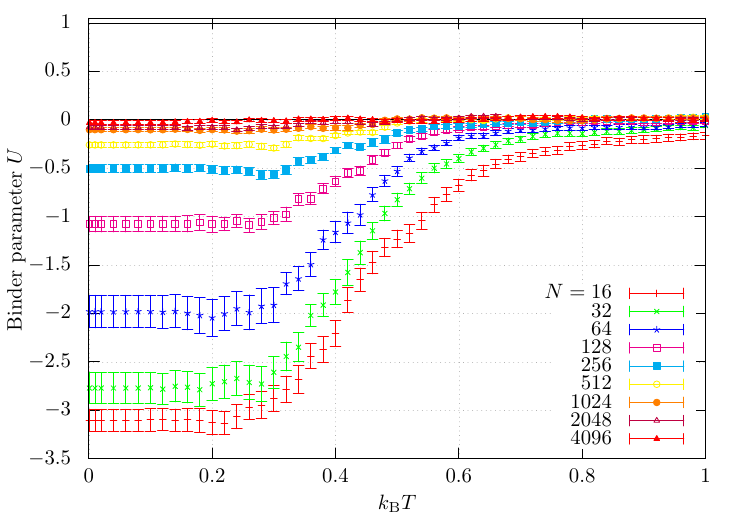}
	\caption{Temperature dependence of the Binder parameter, Eq.~\eqref{main_eq_Binder_parameter_001_001}. The first panel is the result for $t=0.3$, the second panel is the result for $t=0.5$, and the third panel is the result for $t=0.7$. We set $K = 20$, $J = 1.0$, $q = 10^{-1.0}$, and $\epsilon=0.00$. We show the results for various system sizes $N = 16, 32, \dots, 4096$, and the curves for different $N$ are overlaid in each panel.}
	\label{appendix_fig_Binder_K = 20_R=1_q_1.0_e=0.00_t!=0.0_001_001}
\end{figure}

Below, we compare the results for $t = 0.3, 0.5, 0.7$ and for $q=10^{-2.0}, 10^{-1.0}$.
The transition temperatures are estimated to be those values shown in the Table~\ref{appendix_table_critical_temperature_K_20_001_001}.
\begin{table}[t]
	\caption{\label{appendix_table_critical_temperature_K_20_001_001}
		The critical temperature of each parameter set.
		The symbol `-' means no phase transitions occur in the parameter setting.}
	\begin{ruledtabular}
		\begin{tabular}{rccc}
			              & $t=0.3$ & $t=0.5$ & $t=0.7$  \\
			\hline
			$q=10^{-2.0}$ & 0.24    & 0.18    & \text{-} \\
			$q=10^{-1.0}$ & 0.30    & 0.24    & \text{-} \\
		\end{tabular}
	\end{ruledtabular}
\end{table}
The magnetization is shown in Figs.~\ref{appendix_fig_mag_K = 20_R=1_q_2.0_e=0.00_t!=0.0_001_001} and
\ref{appendix_fig_mag_K = 20_R=1_q_1.0_e=0.00_t!=0.0_001_001},
the susceptibility is shown in Figs.~\ref{appendix_fig_sus_K = 20_R=1_q_2.0_e=0.00_t!=0.0_001_001} and
\ref{appendix_fig_sus_K = 20_R=1_q_1.0_e=0.00_t!=0.0_001_001},
and the Binder parameter is shown in Figs.~\ref{appendix_fig_Binder_K = 20_R=1_q_2.0_e=0.00_t!=0.0_001_001} and
\ref{appendix_fig_Binder_K = 20_R=1_q_1.0_e=0.00_t!=0.0_001_001}.


\subsubsection{$k_{\mathrm{B}} T$--$t$ phase diagram}

\begin{figure}[t]
	\centering
	\includegraphics[scale=0.60]{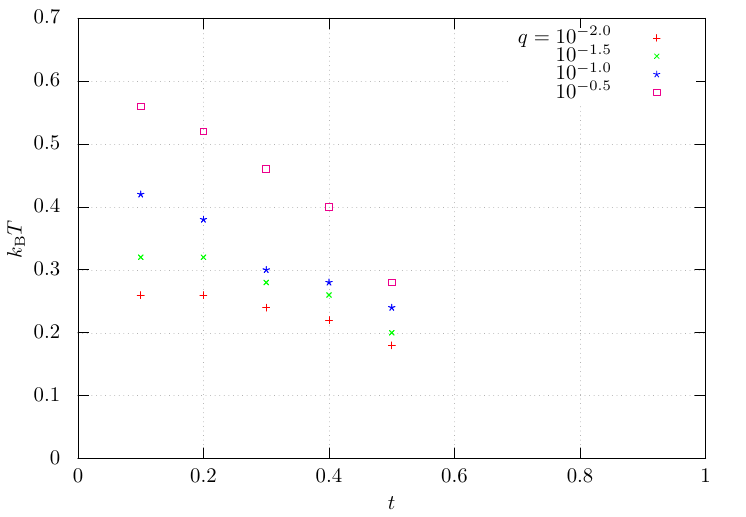}
	\caption{Phase diagram for $K = 20$, $J=1.0$, and $\epsilon=0.00$. The horizontal and vertical axes represent the parameter $t$, which controls the probability of selecting production rules, and the critical temperature $k_{\mathrm{B}} T$, respectively.}
	\label{appendix_fig_t_phase_diagram_K = 20_R=1_e=0.00_001_001}
\end{figure}

The $k_{\mathrm{B}} T$--$t$ phase diagram is shown in Fig.~\ref{appendix_fig_t_phase_diagram_K = 20_R=1_e=0.00_001_001}.

\clearpage

\FloatBarrier
\bibliography{paper_language-model_short_999_001}

\end{document}